\documentclass[10pt,journal,compsoc]{IEEEtran}
\usepackage[nocompress]{cite}
\ifCLASSINFOpdf
\else
\fi
\usepackage{bm}
\usepackage{amsmath}

\usepackage{mdwmath}
\usepackage{eqparbox}
\usepackage{epsfig}
\usepackage{graphicx}
\usepackage{amsmath}
\usepackage{amssymb}
\usepackage{pifont}
\usepackage{url}            % simple URL typesetting
\usepackage{booktabs}       % professional-quality tables
\usepackage{tabu}           % tabular
\usepackage{multirow}
\usepackage{graphicx}
\usepackage{algorithm}
\usepackage{algorithmicx}
\usepackage{algpseudocode}
\usepackage{amsmath,amssymb}
\usepackage[table,dvipsnames]{xcolor}
\usepackage{array}
\usepackage[pagebackref=true,breaklinks=true,colorlinks,bookmarks=false]{hyperref}
\usepackage{stfloats}
\usepackage{cite}

\usepackage{makecell}
\usepackage{tabularx}
\usepackage{multirow}
\usepackage{pifont}
\usepackage{multicol}
\usepackage{array}
\usepackage{adjustbox}

\usepackage{url}            % simple URL typesetting
\usepackage{booktabs}       % professional-quality tables
\usepackage{amsfonts}       % blackboard math symbols
\usepackage{nicefrac}       % compact symbols for 1/2, etc.
\usepackage{microtype}      % microtypography
\usepackage{xcolor}         % colors
\usepackage{graphicx}
\usepackage{amsmath}
\usepackage{bm}
\usepackage{multirow}
\usepackage{tabu}
\usepackage{siunitx}
\usepackage{adjustbox}
\usepackage{subcaption}
\usepackage{siunitx}
\usepackage{xcolor}
\usepackage{colortbl}
\usepackage{color}
\usepackage{pifont}
\usepackage{colortbl}
\definecolor{Gray}{gray}{0.95}
\definecolor{Cyan}{rgb}{0.88,1,1}

\usepackage{tabu}           % tabular
\usepackage{multirow}
\usepackage{booktabs}
\usepackage{makecell}

\newcommand{\paragrapha}[2][3pt]{\vspace{#1}\noindent\textbf{#2}}

\newcolumntype{x}[1]{>{\centering\arraybackslash}p{#1pt}}
\newlength\savewidth

\newcommand{\PreserveBackslash}[1]{\let\temp=\\#1\let\\=\temp}
\newcolumntype{C}[1]{>{\PreserveBackslash\centering}p{#1}}
\newcolumntype{L}[1]{>{\PreserveBackslash\raggedright}p{#1}}

\usepackage{soul}
\sethlcolor{Gray}

\begin{document}
%
% paper title
% Titles are generally capitalized except for words such as a, an, and, as,
% at, but, by, for, in, nor, of, on, or, the, to and up, which are usually
% not capitalized unless they are the first or last word of the title.
% Linebreaks \\ can be used within to get better formatting as desired.
% Do not put math or special symbols in the title.
% \title{Generic Visual Perception with High-Order Spatial Interactions}

% \title{High-Order Recursive Gated Convolution \\ for Generic Visual Perception}
% \title{Generic Visual Perception with Efficient High-Order Spatial Interactions}
\title{Insight-V++: Towards Advanced Long-Chain Visual Reasoning with Multimodal Large Language Models}
%
%
% author names and IEEE memberships
% note positions of commas and nonbreaking spaces ( ~ ) LaTeX will not break
% a structure at a ~ so this keeps an author's name from being broken across
% two lines.
% use \thanks{} to gain access to the first footnote area
% a separate \thanks must be used for each paragraph as LaTeX2e's \thanks
% was not built to handle multiple paragraphs
%
%
%\IEEEcompsocitemizethanks is a special \thanks that produces the bulleted
% lists the Computer Society journals use for "first footnote" author
% affiliations. Use \IEEEcompsocthanksitem which works much like \item
% for each affiliation group. When not in compsoc mode,
% \IEEEcompsocitemizethanks becomes like \thanks and
% \IEEEcompsocthanksitem becomes a line break with idention. This
% facilitates dual compilation, although admittedly the differences in the
% desired content of \author between the different types of papers makes a
% one-size-fits-all approach a daunting prospect. For instance, compsoc
% journal papers have the author affiliations above the "Manuscript
% received ..."  text while in non-compsoc journals this is reversed. Sigh.

\author{Yuhao Dong,~
        Zuyan Liu,~
        Shulin Tian,~
        Yongming Rao,~
        Ziwei Liu
\IEEEcompsocitemizethanks{
\IEEEcompsocthanksitem The authors are with S-Lab, Nanyang Technological University, Tencent Hunyuan, and Tsinghua University. \\
Email: yuhao013@e.ntu.edu.sg; ziwei.liu@ntu.edu.sg. Ziwei Liu is the corresponding author. 
}% <-this % stops a space
}

% \author{Yuhao Dong~\IEEEmembership{Student Member,~IEEE},
%         Zuyan Liu~\IEEEmembership{Student Member,~IEEE},
%         Shulin Tian~\IEEEmembership{Student Member,~IEEE},
%         Yongming Rao~\IEEEmembership{Member,~IEEE},
%         Ziwei Liu~\IEEEmembership{Member,~IEEE}
% \IEEEcompsocitemizethanks{
% \IEEEcompsocthanksitem The authors are with S-Lab, Nanyang Technological University, Tencent Hunyuan, and Tsinghua University. \\
% Email: yuhao013@e.ntu.edu.sg; ziwei.liu@ntu.edu.sg. Ziwei Liu is the corresponding author. 
% }% <-this % stops a space
% }

% The paper headers
\markboth{}%
{}
\IEEEtitleabstractindextext{%
\begin{abstract}

Large Language Models (LLMs) have achieved remarkable reliability and advanced capabilities through extended test-time reasoning, evolving from foundational Chain-of-Thought prompting to sophisticated paradigms like OpenAI o1. However, extending these capabilities to Multi-modal Large Language Models (MLLMs) remains a significant challenge due to a critical scarcity of high-quality, long-chain reasoning data and the absence of optimized training pipelines. To bridge this gap, we present a unified multi-agent visual reasoning framework that systematically evolves from our foundational image-centric model, Insight-V, into a generalized spatial-temporal architecture, Insight-V++. We first propose a scalable, progressive data generation pipeline equipped with multi-granularity assessment that autonomously synthesizes structured, complex reasoning trajectories across both the image and video domains without human intervention. Recognizing that directly supervising MLLMs with such intricate data often yields sub-optimal results, we design a dual-agent architecture. This system comprises a reasoning agent dedicated to executing extensive analytical chains, and a summary agent trained to critically evaluate and distill the final outcomes. While our initial Insight-V framework utilized an iterative Direct Preference Optimization (DPO) algorithm to stabilize generation, the off-policy nature of DPO fundamentally constrained its reinforcement learning potential. To overcome these limitations, particularly for long-horizon video understanding, Insight-V++ introduces two novel reinforcement learning algorithms named ST-GRPO and J-GRPO. These algorithms specifically enhance the spatial-temporal reasoning capabilities of the reasoning agent and fundamentally improve the evaluative robustness of the summary agent. Crucially, we achieve a tighter integration of these modules through a novel self-evolving training strategy, resulting in a highly compact and efficient system. By leveraging reliable feedback signals from the well-trained summary agent, we guide an iterative reasoning path generation process. This enables the system to autonomously produce superior, refined data, which is subsequently utilized to retrain the entire multi-agent system in a continuous, self-improving loop. Extensive experiments built upon robust base models, including LLaVA-NeXT and Qwen2.5-VL, demonstrate that our cohesive framework achieves significant, consistent performance gains across challenging image and video reasoning benchmarks while preserving strong capabilities on traditional perception-focused tasks.
\end{abstract}

% Note that keywords are not normally used for peerreview papers.
}

% make the title area
\maketitle

% To allow for easy dual compilation without having to reenter the
% abstract/keywords data, the \IEEEtitleabstractindextext text will
% not be used in maketitle, but will appear (i.e., to be "transported")
% here as \IEEEdisplaynontitleabstractindextext when compsoc mode
% is not selected <OR> if conference mode is selected - because compsoc
% conference papers position the abstract like regular (non-compsoc)
% papers do!
\IEEEdisplaynontitleabstractindextext
% \IEEEdisplaynontitleabstractindextext has no effect when using
% compsoc under a non-conference mode.

% For peer review papers, you can put extra information on the cover
% page as needed:
% \ifCLASSOPTIONpeerreview
% \begin{center} \bfseries EDICS Category: 3-BBND \end{center}
% \fi
%
% For peerreview papers, this IEEEtran command inserts a page break and
% creates the second title. It will be ignored for other modes.
\IEEEpeerreviewmaketitle

\ifCLASSOPTIONcompsoc
\IEEEraisesectionheading{\section{Introduction}\label{sec:introduction}}
\else
\section{Introduction}
\label{sec:intro}
\fi

\IEEEPARstart{T}{he} development of artificial general intelligence requires models that can seamlessly understand and respond to multi-modal data. Recent advancements in Large Language Models (LLMs)~\cite{GPT4o,qwen2,dubey2024llama,qwen2.5} and Multi-modal LLMs (MLLMs)~\cite{liu2024llava,liu2024llava15,liu2024llavanext,chen2024internvl,qwen2vl,lu2024deepseek,yao2024minicpmv} have significantly facilitated this progress across various fields, ranging from common question-answering~\cite{qwen2vl,chen2024internvl,liu2024llavanext,li2024llavaov,liu2024oryx} to autonomous driving~\cite{tian2024drivevlm,ma2023dolphins} and robotics~\cite{yang2023octopus,driess2023palm}. Despite the substantial progress made in enhancing the performance of MLLMs on a wide range of tasks, enabling MLLMs to perform human-level reasoning remains a key challenge. This area remains underexplored and has yet to fully realize its potential.

\begin{figure*}[t]
\centering
\includegraphics[width=\textwidth]{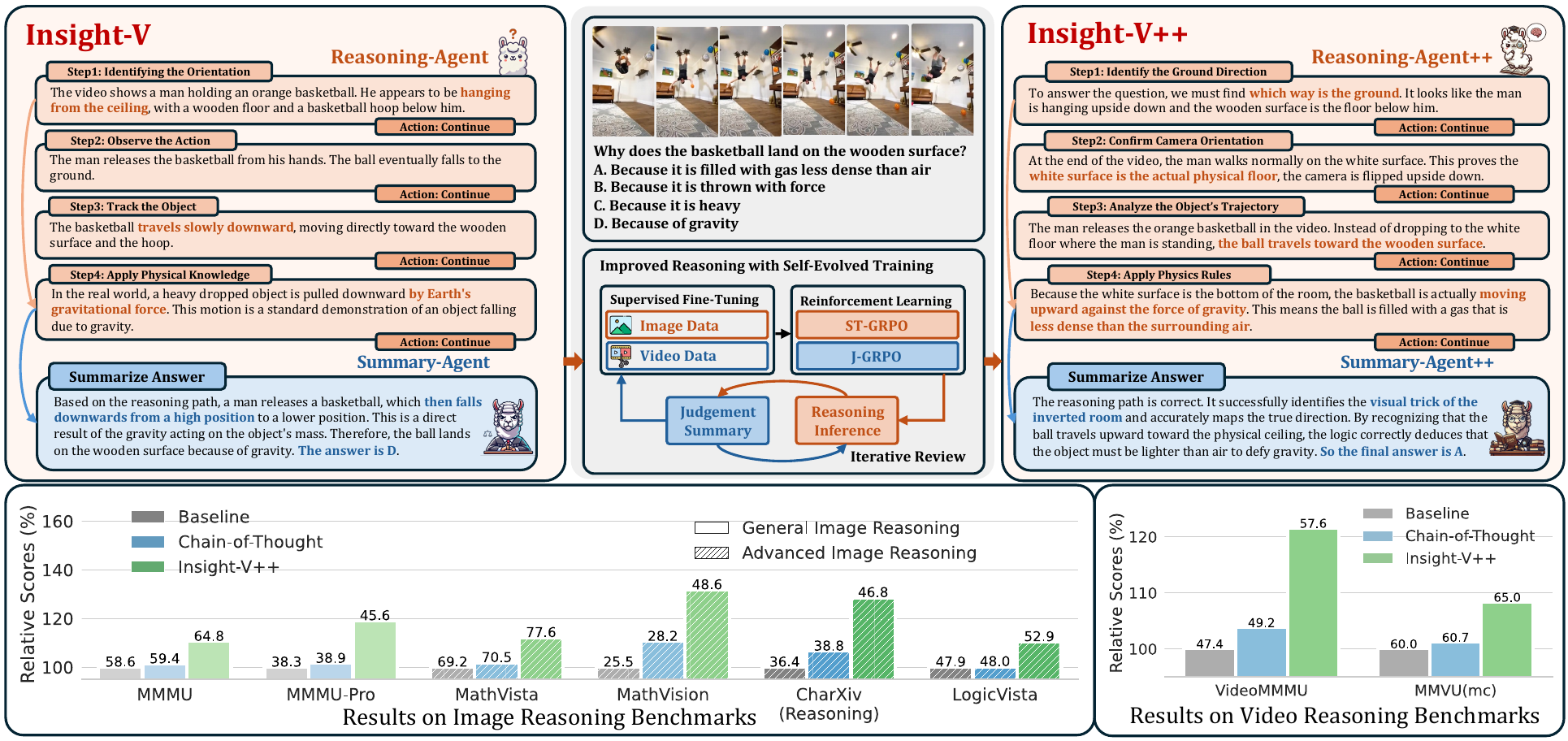}
\caption{\small \textbf{Illustration and Performance of Insight-V and Insight-V++.} Insight-V employs a dual-agent architecture, comprising dedicated reasoning and summarization modules to achieve significant performance gains across various image reasoning benchmarks. Building upon this foundation, Insight-V++ introduces a novel self-evolving training paradigm. By utilizing an iterative loop of Supervised Fine-Tuning (SFT) and Reinforcement Learning (RL) to continuously refine its visual reasoning capabilities, Insight-V++ yields superior results on both image and video benchmarks.}
\label{fig:teaser}
\vspace{-10pt}
\end{figure*}

Existing efforts~\cite{wei2022chain,yao2024tree} to enhance the reasoning capabilities of LLMs through long-chain reasoning have demonstrated considerable progress, largely benefiting from the availability of structured, high-quality data and well-established training pipelines. In contrast, teaching MLLMs to perform long-chain visual reasoning remains a significant challenge, primarily due to the lack of large-scale, high-quality datasets and efficient and effective training strategies. Compared to text-only data, visual reasoning data is not only more expensive to collect but also requires significant human labor for detailed annotation and validation, due to the absence of an effective data generation pipeline. Moreover, while previous work~\cite{zhang2023multimodal} has demonstrated that directly applying chain-of-thought~\cite{wei2022chain} reasoning can improve the capabilities of MLLMs, other research~\cite{zhang2024mavis,zhang2024improve} suggests that current training approaches have limited effectiveness in enhancing CoT reasoning. This highlights the inability of current MLLMs to leverage visual cues for precise step-by-step problem-solving, emphasizing the need for an effective training procedure that enables MLLMs to reason in detail while maintaining clear visual perception.

While establishing this detailed reasoning in static images is a critical first step, the rapid evolution of real-world applications increasingly demands that models extend these capabilities to dynamic environments. Consequently, mastering video reasoning has emerged as a crucial next frontier for multi-modal understanding. However, transitioning from static visual perception to dynamic, long-form video domain introduces a profoundly higher level of complexity. Unlike static images, video reasoning inherently requires models to track shifting objects over time, comprehend intricate action sequences, and maintain rigorous spatial-temporal coherence across numerous frames. As a result, existing data generation pipelines and conventional training strategies are largely inadequate for capturing these dynamic nuances. Furthermore, beyond the specific challenges of temporal modeling, a broader bottleneck restricts the ceiling of generalized reasoning across \textit{both} image and video modalities: the inherent limitations of static, non-adaptive training paradigms. To truly fulfill the need for an effective, scalable training procedure, an architecture must transcend loosely coupled, one-off optimization pipelines. It must instead cultivate the capacity for continuous self-evolution, leveraging reliable internal feedback to autonomously correct, refine, and scale its reasoning capabilities without being restricted by fixed, human-annotated datasets.

To address these challenges, we propose Insight-V, which incorporates two innovative designs to enhance reasoning capabilities. First, we introduce a data generation pipeline consisting of two key steps: a progressive strategy to generate structured, long-chain reasoning data with diverse reasoning paths, and a multi-granularity assessment system to evaluate and score these paths at different levels. Through automatic generation, assessment, and ranking strategies, the pipeline effectively operates without the need for human labor and makes the reasoning dataset more scalable for enhancing reasoning capabilities. To further improve MLLM reasoning beyond data scaling, we design a multi-agent system, as illustrated in Figure~\ref{fig:teaser}, that decomposes the problem-solving process into two distinct steps: reasoning and summarization. The reasoning agent generates a detailed reasoning process for the input query, while the summarization agent identifies key information within the reasoning process and selectively answers the question. To refine the quality of the reasoning, we employ an iterative DPO approach to enhance reasoning capabilities. The two agents collaborate to further improve the reasoning quality. Our findings demonstrate that this system significantly enhances the performance of various MLLMs across a broad range of visual reasoning benchmarks.

To further address the spatial-temporal challenges and elevate our architecture from a static pipeline into a dynamic, self-evolving framework, we introduce Insight-V++. This extension fundamentally unifies image and video domains, addressing the critical limitations of existing models. First, we upgrade our progressive data generation pipeline to autonomously construct complex, step-by-step video reasoning datasets. By introducing an advanced in-context reasoning path scoring method, we rigorously evaluate dynamic trajectories while preserving essential reasoning diversity. Crucially, to resolve the instability and limitations of off-policy DPO in RL training, Insight-V++ shifts to a robust on-policy reinforcement learning paradigm via two novel objectives: ST-GRPO and J-GRPO. By introducing novel reward designs and training strategies, these algorithms ensure highly stable optimization across both image and video data. ST-GRPO effectively forces the reasoning agent to master temporal alignment and complex spatial-temporal logic, while J-GRPO significantly fortifies the summary agent's evaluative robustness. Beyond these algorithmic advancements, the core innovation of Insight-V++ lies in its transformation into a tightly integrated, self-evolving ecosystem. Rather than functioning as isolated modules, the two agents are deeply integrated through a collaborative self-evolving mechanism. By leveraging rigorous feedback signals from the enhanced summary agent, the reasoning agent autonomously corrects, refines, and synthesizes increasingly superior reasoning trajectories. These high-quality trajectories are subsequently fed back into the training pipeline to jointly re-optimize both agents. This continuous self-improvement loop facilitates profound knowledge consolidation and progressively scales the system's reasoning capacity, yielding state-of-the-art spatial-temporal understanding without the need for additional human-annotated data.

Extensive evaluations demonstrate the efficacy of our unified framework across different stages of its evolution. For our foundational Insight-V, integrating the system into the widely used LLaVA-NeXT~\cite{liu2024llavanext} architecture yields an average performance improvement of 8.1\% across six challenging visual reasoning benchmarks. Furthermore, applying it to a stronger base MLLM results in a 3.3\% gain, underscoring its broad generalizability. Building on this success, we instantiate the advanced Insight-V++ framework using the Qwen2.5-VL~\cite{bai2025qwen25vl} model, a baseline model with robust foundational capabilities in both image and video modalities. To rigorously assess Insight-V++, we expand our evaluation suite to encompass highly demanding image reasoning benchmarks alongside comprehensive video reasoning benchmarks. Across both modalities, the framework achieves state-of-the-art results, successfully preserving core perception skills while mastering complex spatial-temporal logic. Notably, on established general image reasoning benchmarks, Insight-V++ secures an additional +4.8\% improvement, a significant gain given the strong Qwen2.5-VL baseline. When subjected to more demanding, high-complexity image tasks, the framework achieves an impressive average score of 53.9, substantially outperforming all previous models built upon this base architecture. Furthermore, in the temporal domain, Insight-V++ delivers an outstanding average improvement of +6.9\% across six representative video reasoning benchmarks. This surpasses existing baselines by a significant margin, directly validating the effectiveness of our spatial-temporal enhancements. Together, these empirical results confirm the strength of the core multi-agent architecture across both models, while explicitly highlighting how the GRPO-based reinforcement learning algorithms and the closed-loop, self-evolving strategy empower Insight-V++ to achieve superior visual reasoning capabilities.

In summary, Insight-V offers \textbf{1)} a scalable data generation pipeline for long-chain, high-quality reasoning data, \textbf{2)} a multi-agent system that decomposes visual reasoning tasks into reasoning and summarization, and \textbf{3)} a two-stage training pipeline to enhance visual reasoning capabilities. 

As an extended version of our previous conference work~\cite{dong2025insight}, Insight-V++ fundamentally unifies and significantly advances our architecture with the following new contributions:
\textbf{4) Unified Spatial-Temporal Data Pipeline:} We seamlessly adapt our progressive generation strategy to the video domain. By proposing a novel in-context scoring mechanism, the framework automatically curates large-scale, high-fidelity dynamic trajectories while maintaining rich structural diversity.
\textbf{5) On-Policy Reinforcement Learning for Video:} To overcome the instability of off-policy optimization, we design ST-GRPO and J-GRPO. These tailored algorithms provide stable reward signals that explicitly equip the reasoning agent with superior spatial-temporal alignment and enhance the summary agent's judgment reliability.
\textbf{6) Closed-Loop Self-Evolution:} We elevate the multi-agent system into a fully collaborative ecosystem. By utilizing evaluation feedback from the summary agent to iteratively generate and filter superior reasoning paths, we establish a continuous co-optimization loop that progressively scales the framework's spatial-temporal capabilities entirely without additional human annotation.

Together, these advancements establish Insight-V++ as a comprehensive, self-improving ecosystem that fundamentally bridges the gap between static image perception and complex spatial-temporal understanding. By autonomously overcoming the multi-modal data scarcity bottleneck and stabilizing long-horizon reasoning through robust on-policy reinforcement learning, our integrated framework paves a highly scalable, annotation-free pathway toward developing the next generation of general-purpose visual reasoning models.

\vspace{-10pt}
\section{Related Work}
\label{sec:related}

\paragrapha{Vision-Language Reasoning.} Recent advances in MLLMs~\cite{liu2024llava,liu2024llava15,liu2024llavanext,lin2023vila,bai2023qwenvl,lu2024deepseek,qwen2vl,liu2024oryx,li2024llavaov} have endowed these models with strong reasoning abilities across domains such as visual understanding~\cite{lin2023vila,qwen2vl}, mathematics~\cite{liang2023unimath}, and scientific problem-solving~\cite{chen2024internvl}. 
In visual understanding, research~\cite{liu2024llavanext,li2023monkey,tong2024cambrian,xu2024llavauhd,liu2024chain} focuses on fine-grained detail analysis and localization, enabling models to perform more interpretable visual reasoning.
Video reasoning benchmarks have evolved from basic question answering~\cite{yu2019activitynet,xiao2021nextqa} to multi-scale evaluations~\cite{fu2024videomme,li2024mvbench,liu2024tempcompass,wu2024longvideobench} that assess temporal reasoning across diverse durations. Recent benchmarks~\cite{hu2025videommmu,zhao2025mmvu,song2025videommlu} further target expert-level and scientific reasoning, while others~\cite{qi2025vcrbench,cheng2025videoholmes,zhang2025videott,demoicl} examine chain-of-thought and multi-clue integration, revealing that current MLLMs still lag behind humans in complex multi-step reasoning.
For mathematical and expert-level reasoning, existing studies~\cite{gao2023g,zhang2024mavis,zhang2024improve} build on Chain-of-Thought (CoT)~\cite{wei2022chain} methods to produce step-by-step solutions. 
Recent works~\cite{xu2025llava,dong2025insightv,yao2024mulberry} extend CoT with multistage reasoning, including summarization, interpretation, and conclusion, demonstrating that structured reasoning paths enhance performance. Other efforts~\cite{guo2025mammothvl,zhang2025openmmreasoner} construct large-scale multimodal instruction-tuning datasets with intermediate rationales, emphasizing the importance of data quality.
However, most methods prioritize dataset quality over structured multistage reasoning, and single-model reasoning remains limited. To address this, we propose a scalable reasoning data pipeline and a multi-agent framework that decomposes reasoning and summarization to enhance MLLM reasoning capabilities.

\paragrapha{Vision-Language Alignment.} To better align MLLMs with human intent, most methods adopt Reinforcement Learning from Human Feedback (RLHF)~\cite{bai2022training} or Direct Preference Optimization (DPO)~\cite{rafailov2024direct}, which directly optimizes human-labeled preference pairs without a reward model. However, conventional DPO is typically offline and can degrade as models evolve. Iterative DPO~\cite{chen2024self} addresses this by repeatedly generating and refining preference pairs; we adopt this strategy to strengthen preference alignment and reasoning ability.
Beyond preference-based alignment, Group Relative Policy Optimization (GRPO)~\cite{shao2024deepseekmath} improves RL efficiency by normalizing rewards within groups, removing the need for a critic network. DeepSeek-R1~\cite{guo2025deepseek} demonstrates that complex reasoning can emerge from GRPO alone, while subsequent works~\cite{liu2024understanding,yu2025dapo,zheng2025group,liu2026gdpo} enhance its stability at scale.
Recent studies extend GRPO-based RL to vision-language models using verifiable visual rewards~\cite{meng2025mmeureka,shen2025vlmr1,chen2025sft}. Yet, purely RL-based training struggles to induce higher-order reasoning, motivating hybrid pipelines that first fine-tune and then apply GRPO~\cite{huang2025visionr1,peng2025lmmr1,wei2025ovr,zhang2025openmmreasoner}, or alternate between supervised and RL stages with progressively harder data~\cite{deng2025openvlthinker}. For the Insight-V series, we integrate advanced reinforcement learning strategies to boost overall performance. Specifically, we propose ST-GRPO and J-GRPO, which are tailored to enhance the distinct capabilities of the reasoning and summary agents, thereby enhancing the overall visual reasoning capabilities of the multi-agent system.

\paragrapha{Agentic Visual Reasoning.}Early work demonstrates that LLMs and MLLMs can orchestrate external vision tools through generated programs or structured prompts to solve compositional visual tasks~\cite{gupta2023visprog,suris2023vipergpt,yang2023mmreact,lu2023chameleon,liu2024llavaplus,hu2024visualsketchpad,su2025openthinkimg}. In the video domain, agentic frameworks~\cite{fan2024videoagent,wang2024videoagent,yuan2025videodeepresearch,zhang2024omagent,tian2025egor1,yang2025longvt} employ LLMs as central controllers that iteratively search, retrieve, and reason over video segments. However, single-agent reasoning is prone to error compounding across intermediate steps~\cite{huang2023large,tyen2024llms}, and self-refinement~\cite{madaan2023selfrefine,shinn2023reflexion} or external verifiers~\cite{cobbe2021training,hosseini2024vstar,lightman2024lets} remain bounded by the individual model's capabilities. Multi-agent debate and collaborative frameworks~\cite{du2024debate,hong2023metagpt,wu2024autogen} address this by distributing reasoning across communicating agents, while role-specialized multi-agent reinforcement learning~\cite{wan2025rema,zhang2025dramamr} further decomposes reasoning into hierarchical agents jointly optimized via multi-turn GRPO. In the multimodal setting, recent work decomposes visual reasoning into functionally distinct agents. MACT~\cite{yu2025mact} assigns planning, execution, judgment, and answering to separate VLMs with adaptive test-time scaling. InSight-o3~\cite{li2025insighto3} decouples reasoning from generalized visual search via a vReasoner-vSearcher pair, and Critic-V~\cite{zhang2025criticv} introduces a Reasoner-Critic architecture inspired by the Actor-Critic paradigm, yet all rely on prompt-based coordination or separately trained components without joint optimization. In the Insight-V series, we introduce a multi-agent framework that decouples complex tasks into specialized reasoning and summary agents. We further leverage a self-evolving training paradigm, which enables the system to achieve tightly coupled collaboration, driving robust visual reasoning across both image and video domains.

\section{Method}
\label{sec:method}

In this section, we provide a comprehensive description of the proposed Insight-V and Insight-V++ system, detailing its architecture and key contributions. Section~\ref{sec:overview} presents an overview of Insight-V and Insight-V++, highlighting the core concepts of our approach. The design of the multi-agent system is structured around three primary components: \textbf{1)} a carefully constructed pipeline for structured, scalable reasoning data generation, as described in Section~\ref{sec:data}; \textbf{2)} a multi-agent MLLM system that supports complex visual reasoning, as detailed in Sections~\ref{sec:model}; and \textbf{3)} a streamlined yet effective training pipeline to enhance overall performance, as outlined in Section~\ref{sec:implementation} and Section~\ref{sec:evolve}. Together, these components form a cohesive system that effectively tackles the challenges of performing detailed, long-chain image and video reasoning while preserving visual perception capabilities.

\subsection{Overview}
\label{sec:overview}
Advancing reasoning capabilities of LLMs has been a focal point of extensive research. Despite these efforts, the reasoning potential within multi-modal LLMs has barely been explored. Most approaches aim to strengthen reasoning at the inference stage, assuming the model has already acquired robust reasoning skills. Other approaches optimize model parameters using chain-of-thought data, enabling models to mimic human reasoning processes. However, these methods present significant challenges for current general-purpose MLLMs, as they require the model to develop reasoning skills while retaining prior capabilities, which often results in only modest performance gains. Additionally, the lack of structured, high-quality training data impedes training models with advanced reasoning capabilities.

To fully leverage the reasoning capabilities of MLLMs, we propose Insight-V, a novel system comprising two MLLMs dedicated to reasoning and summarization, respectively. The \texttt{reasoning model} is tasked with generating a detailed reasoning process to assist in problem-solving, while the \texttt{summary model} evaluates this reasoning as supplementary information to assess its relevance and utility for answering the question. We also construct a structured, high-quality dataset to train both agents. We posit that this multi-agent system can enhance the reasoning strengths of MLLMs by decomposing the problem-solving process into distinct reasoning and summarization phases, thereby driving substantial performance improvements.

To further enhance general visual reasoning capabilities, we introduce Insight‑V++. Specifically, we extend the data generation pipeline to construct high-quality video reasoning datasets using a newly designed in-context scoring strategy. In addition, we propose two reinforcement learning algorithms: ST‑GRPO and J‑GRPO, for training the reasoning agent and summary agent, respectively. To fully exploit the summary agent and make the system more compact and adaptive, we develop a self-evolving strategy in which the reasoning agent iteratively generates improved reasoning paths with feedback from the summary agent. The resulting data are then used to fine-tune the initial reasoning agent, thereby enhancing the overall reasoning capability of the system. These design innovations advance the multi-agent framework, making it more general, self-evolving, and well-suited for complex reasoning tasks.

\begin{figure}[t]
\centering
\includegraphics[width=0.5\textwidth]{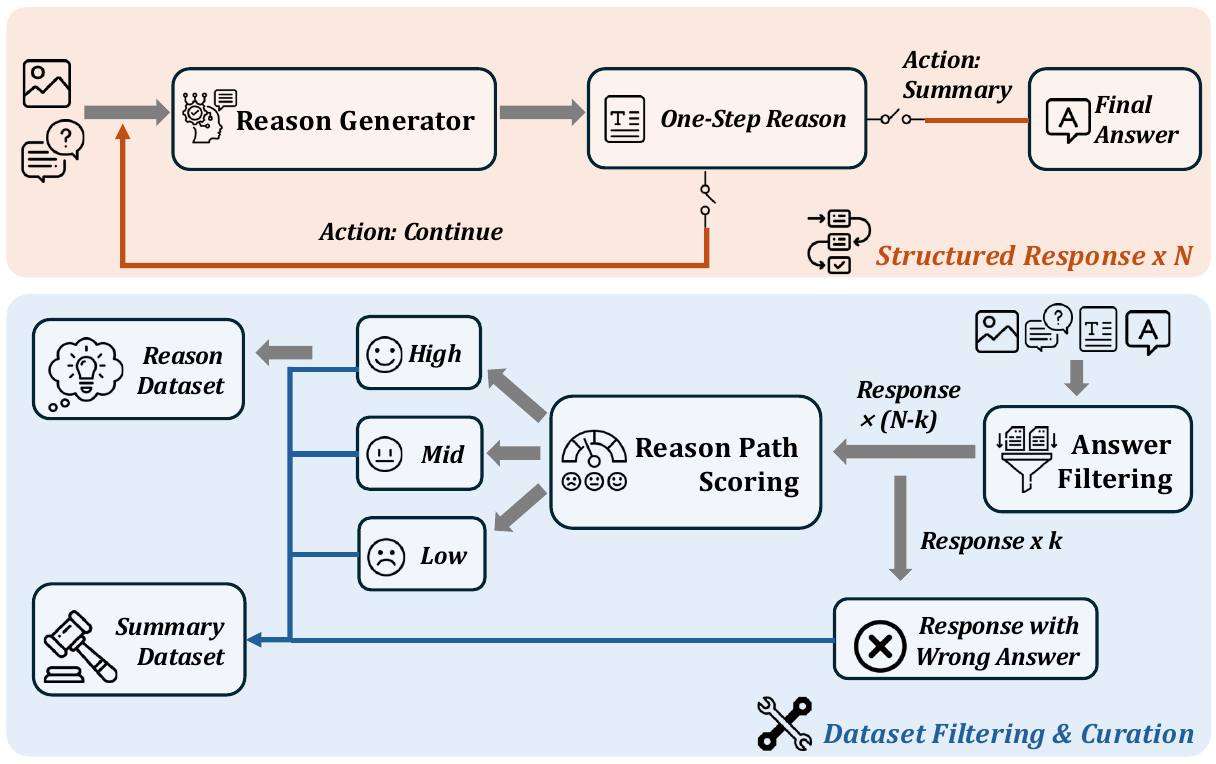} 
\caption{\textbf{Data Generation Pipeline of Insight-V Series. } The reasoning processes are generated progressively through a reasoning generator, and then fed into a multi-granularity assessment system to ensure high-quality reasoning.}
\label{fig:pipeline}
\end{figure}

\subsection{Construction of Structured Reasoning Data}
\label{sec:data}
Previous studies~\cite{zhang2023multimodal,zhang2024mavis} have explored the integration of reasoning capabilities into MLLMs. However, training MLLMs to develop robust reasoning skills remains a considerable challenge, particularly due to data limitations. To address this, we introduce our data generation pipeline in this section, designed to produce high-quality, long-chain reasoning data using a progressive generation process and multi-granularity assessment. As shown in Figure~\ref{fig:pipeline}, this scalable approach enables us to generate high-quality data to enhance the model's reasoning capabilities effectively.

\paragrapha{Progressive Long-Chain Reasoning Data Generation. } For each input query, we first employ a reasoning generator to produce a structured reasoning process in JSON format to address the problem. At each step, the reasoning generator provides a brief summary of the current step, a detailed reasoning response, and an action for the following step. If the action is \( continue \), the model proceeds with an additional reasoning step in the next iteration; if the action is \( summary \), the model generates a final summary and answer based on the complete reasoning process in the subsequent iteration. Specifically, For a multi-modal model \( M \), input image \( I \), and question \( Q \), the data generation process for each step is represented as follows:
\begin{align*}
&R_{t}  =   M(I, Q, [R_{1} \cdots R_{t-1}], A), \\
&R_{ans}  = M(I, Q, [R_{1} \cdots R_{n}]),
\end{align*}
where \( R_{t} \) and \( R_{ans} \) denote the response at the \( t \)-th step and the final answer, respectively, \( R_{i} \) represents the reasoning generated by the model at the \( i \)-th step, \(n\) represents the total reasoning steps, and \( A \) is the action determined in the previous step. By repeating this process \( N \) times, we can iteratively sample \( N \) structured responses for each query. The generation parameters are adjusted to encourage the model to produce outputs with various information and steps, allowing us to identify the most effective reasoning chain for each question.

\paragrapha{Multi-Granularity Assessment. } After obtaining the structured responses, we utilize an assessment pipeline to ensure data quality. Specifically, we first apply a strong LLM, such as Qwen2~\cite{qwen2}, for direct \textbf{answer filtering}. As shown in Figure~\ref{fig:pipeline}, the model is provided with the generated final answer and the ground truth answer and is tasked with determining whether the generated answer is correct, serving as an approximate indicator of the validity of the associated reasoning chain. Once responses with incorrect answers are filtered out, the remaining reasoning processes are passed to a \textbf{reasoning path scoring} agent. Here, an advanced multi-modal model, such as Qwen2-VL~\cite{qwen2vl}, is supplied with the image, question, reasoning path, and ground truth answer, and is prompted to evaluate the reasoning path. The scoring agent assesses each response based on the step-by-step accuracy of the reasoning path and the level of detail in the reasoning. To ensure consistency in scoring across different data samples, we aggregate all responses for each question and process them in a single pass. The model then generates scores for each response, ranging from 1 to 100.

Through the above two steps, we construct a structured, high-quality image dataset that provides detailed reasoning for each question, effectively supporting the training of our models.

\paragrapha{Video Reasoning Data Construction. }Video data present greater challenges for current MLLMs due to their inherent temporal dynamics and long-context dependencies, which make it substantially more difficult to produce coherent and high-quality reasoning trajectories. To address this limitation, we extend the image-based data generation pipeline to the video domain and introduce an advanced data scoring strategy based on in-context examples, thereby improving the quality of the generated video reasoning data.

To construct video reasoning data, we first adopt a progressive data generation pipeline to produce long-chain video reasoning trajectories. However, assessing their quality remains more challenging than in image-based visual reasoning, as verifying both the correctness and coherence of reasoning paths over temporal contexts is inherently difficult. To achieve accurate and scalable quality evaluation, we first apply direct answer filtering using Qwen2.5‑72B~\cite{qwen2.5} to ensure the fundamental reliability of the generated data. We then introduce in-context examples created by state-of-the-art models proficient in video reasoning, such as Gemini‑2.5‑Pro. To maintain data diversity, we carefully select representative video questions covering general video understanding, temporal grounding, causal reasoning, and fine-grained video analysis, each verified by human annotators to form a high-quality golden set of video reasoning cases. Next, we feed the video, question, reasoning path, and ground-truth answer, together with the golden in-context examples, into Qwen2.5‑VL~\cite{bai2025qwen25vl}. The model is then prompted to evaluate the quality of each reasoning path based on the provided in-context exemplars, which illustrate what constitutes strong video reasoning behavior. Similar to the image data generation process, all reasoning paths are processed in a single forward pass, yielding quality scores ranging from 1 to 100.

By incorporating golden video reasoning data as in-context examples, we aim to guide the judge model in understanding the characteristics of high-quality video reasoning, as current open-source models still lack this capability. This in turn enhances the reliability and overall quality of the scored video reasoning data.

\begin{figure*}[t]
\centering
\includegraphics[width=\textwidth]{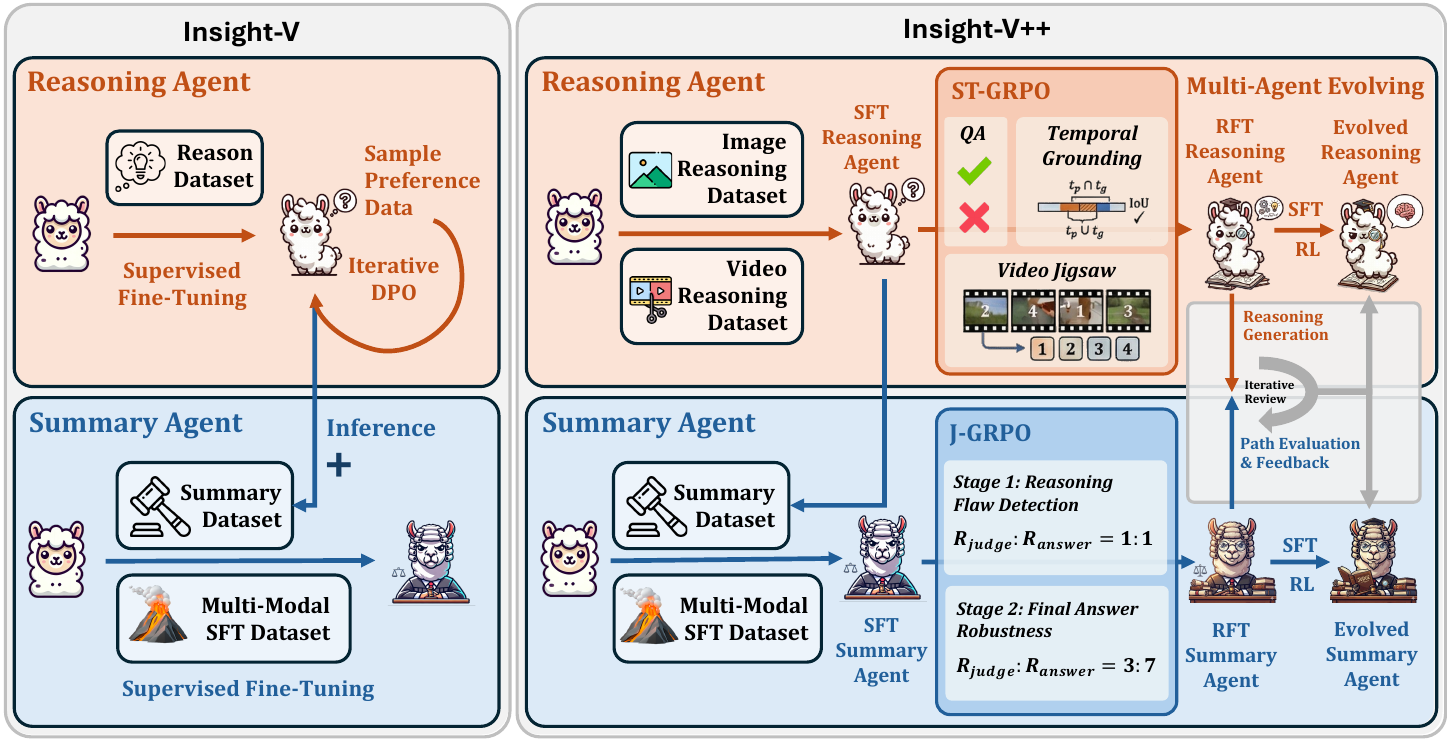} 
\caption{\textbf{Overview of Insight-V and Insight-V++ Model Design.} In Insight-V, we derive a multi-agent system from a single base model. By decomposing the task into reasoning and summarization, these two agents collaborate to enhance overall performance. For Insight-V++, we further introduce ST-GRPO and J-GRPO, which are tailored specifically to each agent. Coupled with a self-evolving strategy, this training loop consisting of SFT and customized RL drives the multi-agent system to excel at visual reasoning.}
\label{fig:method}
\end{figure*}

\subsection{Model Design}
\label{sec:model}
After constructing the dataset, we develop a multi-agent framework to enhance overall reasoning capabilities through collaborative agent interaction. Specifically, we first train a reasoning agent to generate a detailed reasoning process for each problem. Then, a summary agent is employed to answer the question, selectively utilizing the reasoning process based on its assessment. Together, these two agents collaborate to improve reasoning performance effectively.

\paragrapha{Reasoning Agent. } Previous approaches typically combine reasoning and question-answering within a single process, which poses challenges for MLLMs. Generating a long-chain reasoning process can introduce errors, and directly answering questions based on flawed reasoning often leads to poorer results. To address this, we propose a specialized reasoning agent designed to generate a detailed, step-by-step reasoning process in response to an input query. We construct the reasoning dataset by selecting the highest-scoring reasoning path for each question. After training on this dataset, the model transforms into a reasoning agent with enhanced reasoning capabilities, enabling it to generate more detailed, structured reasoning processes.

\paragrapha{Summary Agent. }
Summarization plays a critical role in enabling models to accurately answer questions. After generating multi-step reasoning, summarization provides a cohesive understanding of the reasoning process, ultimately guiding the model to the final answer. However, since the response generated by the reasoning agent may contain errors, we develop a summarization model robust to inaccuracies in the reasoning path, selectively incorporating or disregarding elements as needed. This approach maximizes the reasoning model’s effectiveness while minimizing the risk of introducing misleading information.

To enhance the robustness of the summary agent, we carefully curate its training dataset. We utilize the collected dataset, which is comprised of two types: data with optimal reasoning processes and data with flawed reasoning processes for the summarization task. This method prevents the model from simply copying reasoning outcomes and encourages critical evaluation of reasoning quality. To further promote critical analysis by the summary agent, we select flawed reasoning samples based on their performance scores. Specifically, we draw flawed responses from varying score ranges to create a dataset with different levels of error, prompting the model to assess reasoning processes at various granularities. To better align the summary model with the reasoning agent, we also incorporate question-reasoning pairs generated by the reasoning agent to enhance collaboration between the two agents. Additionally, to preserve the original multi-modal capabilities, we supplement the dataset with standard question-answering data to sustain the summary agent's performance in direct question-answering.

\vspace{-10pt}
\subsection{Multi-Agent Training Pipeline}
\label{sec:implementation}

The training pipeline for Insight-V is designed to be straightforward and efficient, utilizing a two-stage strategy. For both the reasoning agent and the summary agent, we begin with a well-trained MLLM. In the first stage, we apply supervised fine-tuning to train the agents to fulfill their designated roles using corresponding datasets. In the second stage, we implement direct preference optimization, following prior research~\cite{zhang2024improve,xiong2024llavaovchat}. This optimization is applied to the reasoning model, aligning it with human reasoning processes in a simple yet effective manner. These two stages enable the development of a robust system with enhanced visual reasoning capabilities.

\subsubsection{Supervised Fine-tuning for Multi-agent System}
To perform supervised fine-tuning and obtain the two agents, we first train a base multi-modal model, following established methodologies. This model can address general visual question-answering tasks and gain foundational vision-language skills. We compile a high-quality image-text dataset focused on knowledge learning to train the base model. This data is sourced from various open-source academic datasets, including LLaVA-OneVision~\cite{li2024llavaov}, Cauldron~\cite{laurenccon2024matters}, and Cambrian-1~\cite{tong2024cambrian}. Once the base model is trained, we further fine-tune the two agents, initializing them from the base model. Note that for Insight‑V++, since our model is initialized from Qwen2.5‑VL~\cite{bai2025qwen25vl}, the first stage of base model training is unnecessary. For the reasoning agent, we utilize the curated reasoning dataset to develop step-by-step reasoning capabilities. For the summary agent, we formulate a dataset as outlined in Section~\ref{sec:model} and sample about one million general image-text pairs from the dataset used for the base model, preserving its original visual perception abilities. Furthermore, for Insight‑V++, we additionally sample general video understanding data from open‑source datasets such as LLaVA‑Video~\cite{zhang2024video}, Oryx~\cite{liu2024oryx}, and Video-R1~\cite{feng2025videor1} to enhance its video comprehension capacity.

\subsubsection{Enhanced Reasoning with DPO Training}

Preference learning has gained increasing focus in the field of large language models. The primary aim is to fine-tune model outputs to align better with human (or expert) preferences, creating outputs more suited to real-world applications. Let’s assume a preference dataset defined as \(\mathcal{D} = \{(x^{(i)}, y_w^{(i)}, y_l^{(i)})\}_{i=1,\dots,|\mathcal{D}|}\), where each \(x^{(i)}\) is prompt, and \(y_w^{(i)}\) and \(y_l^{(i)}\) represent the preferred and less preferred responses, respectively. We denote \(y_w \succ y_l \mid x\) to signify that \(y_w\) is preferred over \(y_l\) for prompt \(x\).

Since the true distribution of human preferences cannot be directly observed, we approximate it with a latent reward model \(r^*(x, y)\), assuming that higher rewards correspond to stronger preferences. Following the approach by~\cite{rafailov2024direct}, we can model the human preference distribution \(p^*\) using the Bradley-Terry (BT) model~\cite{bradley1952rank}:
\[
\begin{aligned}
p^*(y_1 \succ y_2 \mid x) &= \frac{\exp(r^*(x, y_1))}{\exp(r^*(x, y_1)) + \exp(r^*(x, y_2))} \\
&= \sigma(r^*(x, y_1) - r^*(x, y_2)),
\end{aligned}
\]
where \(\sigma\) denotes the logistic function. 

To estimate the parameters of the reward model, we can apply maximum likelihood estimation by minimizing the negative log-likelihood:
\[
\mathcal{L}_R(r_\phi, \mathcal{D}) = - \mathbb{E}_{(x, y_w, y_l) \sim \mathcal{D}} [\log \sigma(r_\phi(x, y_w) - r_\phi(x, y_l))],
\]
where \(r_\phi\) is a parameterized reward model. This approach allows us to approximate the preference distribution and fine-tune the model to capture human-like preferences effectively.

The traditional DPO algorithm operates in an offline setting. During DPO training, as model parameters continuously change, the preference dataset generated offline can gradually diverge from the model’s current distribution, which weakens the effectiveness of the DPO algorithm. To address this issue, we employ an iterative DPO algorithm. By conducting multiple rounds of DPO training and sampling, this approach enables the model to better approximate an online setting during training, thus further enhancing its performance. Specifically, our approach involves training a sequence of models \( M_1, \ldots, M_T \), where each subsequent model \( M_{t+1} \) utilizes preference data \( \mathcal{D}_t \) generated by the \( t \)-th model. We apply this complete training process to the fine-tuned reasoning agent, enabling the model to better align with human preferences and produce structured, detailed reasoning steps for complex questions, which supports the summary agent more effectively.

\subsubsection{Improved RL training with ST-GRPO and J-GRPO}

\paragrapha{Preliminary.} Group Relative Policy Optimization (GRPO)~\cite{shao2024deepseekmath} is a reinforcement learning algorithm that eliminates the need for a separate critic model by estimating baselines from group-level statistics. Specifically, for each input $q$,  GRPO samples a group of $G$ outputs $\{o_1, o_2, \dots, o_G\}$ from the old policy $\pi_{\theta_{\text{old}}}$ and computes a reward $r_i$ for each output. The advantage of each output is then estimated by normalizing the rewards within the group:
\begin{equation}
    \hat{A}_i = \frac{r_i - \mathrm{mean}(\{r_j\}_{j=1}^{G})}{\mathrm{std}(\{r_j\}_{j=1}^{G})}
\end{equation}
where $\mathrm{mean}(\cdot)$ and $\mathrm{std}(\cdot)$ denote the mean and standard deviation of the group rewards, respectively. This normalization assigns positive advantages to above-average outputs and negative advantages to below-average ones, serving as a baseline estimator that replaces the learned critic in standard actor-critic methods.
The policy $\pi_{\theta}$ is then optimized by maximizing the following objective:
\begin{equation}
\begin{aligned}
    \mathcal{J}_{\text{GRPO}}(\theta) &= \mathbb{E}_{q,\, \{o_i\}_{i=1}^{G}} \Bigg[ \frac{1}{G} \sum_{i=1}^{G} \frac{1}{|o_i|} \sum_{t=1}^{|o_i|} \bigg( \\
    & \quad \min\!\Big( \rho_{i,t}\, \hat{A}_i,\; \mathrm{clip}(\rho_{i,t},\, 1{-}\varepsilon,\, 1{+}\varepsilon)\, \hat{A}_i \Big) \\
    & \quad - \beta\, D_{\mathrm{KL}}\!\left(\pi_\theta \| \pi_{\mathrm{ref}}\right) \bigg) \Bigg],
\end{aligned}
\end{equation}
where $\rho_{i,t} = \frac{\pi_\theta(o_{i,t} \mid q, o_{i,<t})}{\pi_{\theta_{\text{old}}}(o_{i,t} \mid q, o_{i,<t})}$ is the importance sampling ratio between the current and old policies at the $t$-th token of the $i$-th output, and $|o_i|$ denotes the sequence length of output $o_i$. The $\mathrm{clip}(\cdot, 1{-}\varepsilon, 1{+}\varepsilon)$ function constrains the ratio to prevent excessively large policy updates, and the $\min(\cdot)$ operator selects the more conservative estimate to ensure stable optimization. The term $D_{\mathrm{KL}}(\pi_\theta \| \pi_{\mathrm{ref}})$ is the KL divergence between the current policy and a reference policy $\pi_{\mathrm{ref}}$ (typically the initial SFT model), weighted by $\beta$, which regularizes the policy to prevent reward hacking and catastrophic forgetting.

Building upon this foundation, we design two specialized GRPO variants tailored to the distinct roles within our multi-agent system: ST-GRPO and J-GRPO.
By independently optimizing each agent with role-specific GRPO objectives, the two agents develop complementary strengths, resulting in a more capable and robust multi-agent system overall.

\paragrapha{Reward and Training Design.}
To enhance the spatial-temporal reasoning of the reasoning agent and the robustness of the summary agent, we designed two tailored reinforcement learning strategies, ST-GRPO and J-GRPO. For ST-GRPO, we define a composite reward 
$$\mathcal{R} = 0.9 \cdot \mathcal{R}_{task} + 0.1 \cdot \mathcal{R}_{format}$$ where $\mathcal{R}_{task}$ integrates multiple objectives. For standard rule-based question answering, we apply a binary $0/1$ reward based on answer correctness. In temporal grounding tasks, the reward is determined by the Intersection over Union (IoU) between the predicted area ($t_p$) and the ground truth ($t_g$): $$\mathcal{R}_{task} = \frac{t_p \cap t_g}{t_p \cup t_g}$$ Furthermore, to improve temporal perception, we draw inspiration from *Visual Jigsaw* by segmenting videos into $N$ clips and requiring the model to restore their original shuffled sequence. This task employs a simplified reward function defined as $$\mathcal{R}_{task} = \frac{1}{N} \sum_{i=1}^{N} (p_i == g_i)$$
where $p_i$ and $g_i$ represent the $i$-th predicted and ground-truth clips, respectively.

For J-GRPO, we train a summary agent using scored reasoning paths, which we further categorized into five quality levels. Given a prompt and a reasoning path, the agent must evaluate the path's quality ($\mathcal{R}_{judge}$) and deduce the correct answer ($\mathcal{R}_{answer}$). To make the training more effective, we employ a curriculum strategy with adaptive reward weighting, using the path quality levels to indicate difficulty. In the first stage of RL training, we mainly sample high-quality paths and weight $\mathcal{R}_{judge}$ and $\mathcal{R}_{answer}$ equally, prompting the agent to focus on identifying reasoning flaws. After this, in the second stage of RL training, we introduce lower-quality reasoning paths and shift the reward ratio to 3:7 for $\mathcal{R}_{judge}$ and $\mathcal{R}_{answer}$, increasingly prioritizing $\mathcal{R}_{answer}$. This shift reflects a core design choice: while the reasoning agent can effectively use summary feedback to correct minor flaws, severely degraded reasoning paths often prove too challenging for the model to rectify. Therefore, emphasizing the final answer in later stages bolsters the summary agent's robustness and general capabilities. The resulting reward function is defined as:

$$\mathcal{R} = 0.9 \cdot (\alpha \cdot \mathcal{R}_{judge} + (1-\alpha) \cdot \mathcal{R}_{answer}) + 0.1 \cdot \mathcal{R}_{format}$$

Through these dual strategies, we equip the reasoning agent with sophisticated spatial-temporal logic while ensuring the summary agent provides both accurate judgments and robust final answers.

\subsection{Multi-Agent Evolving System}
\label{sec:evolve}
Previous studies~\cite{dong2025insight, zhang2025critic} introduce an auxiliary agent designed primarily to assist the main agent, for example by providing feedback on incorrect reasoning trajectories or enhancing robustness against flawed reasoning paths. However, these designs mainly focus on improving inference-time performance through multi-agent collaboration, while relatively few works explore how to fully leverage the potential of multi-agent systems to enhance the fundamental reasoning capabilities of the agents themselves.

In Insight-V++, we propose a self-evolving training strategy to systematically enhance the model’s overall capabilities. Following the reinforcement learning phase, we acquire two robust, specialized agents: a reasoning agent dedicated to generating reasoning paths, and a summary agent focused on evaluating these paths and providing reliable answers. To fully leverage their complementary strengths, we establish a collaborative reasoning process driven by the original supervised fine-tuning dataset. For each sample, the reasoning agent formulates an initial reasoning trajectory. The summary agent then reviews this path, identifies potential flaws, and provides corrective feedback alongside the final answer. Subsequently, the reasoning agent refines its path conditioned on both its previous output and the newly received feedback. Formally, during the $n$-th iteration of this collaborative process, the outputs of the two agents are defined as:
$$\mathcal{O}_{\mathcal{R}_n} = \mathcal{A}_\mathcal{R}(\mathcal{O}_{\mathcal{R}_{n-1}}, \mathcal{O}_{\mathcal{S}_{n-1}}); \quad \mathcal{O}_{\mathcal{S}_n} = \mathcal{A}_\mathcal{S}(\mathcal{O}_{\mathcal{R}_{n}})$$
where $\mathcal{A}_\mathcal{R}$ and $\mathcal{A}_\mathcal{S}$ denote the reasoning and summary agents, respectively, while $\mathcal{O}_{\mathcal{R}_n}$ and $\mathcal{O}_{\mathcal{S}_n}$ represent their corresponding outputs. This iterative interaction continues until the summary agent deems the reasoning path satisfactory, capped at a maximum of three iterations to ensure efficiency in practice.

This collaborative reasoning process naturally yields a corpus of high-quality reasoning trajectories. To drive our self-evolving strategy, we rigorously filter these generated examples using the data assessment pipeline introduced before and utilize them to further fine-tune the reasoning agent, substantially enhancing its inferential capabilities. Concurrently, we curate this newly acquired data to enrich the training corpus of the summary agent, facilitating its continuous optimization. Finally, we conduct a subsequent reinforcement learning phase using these updated models to yield a more robust, intelligent, and adaptive framework. Through this continuous cycle of generation, evaluation, and targeted refinement, the two agents engage in a mutually beneficial collaboration, enabling the overall system to autonomously self-evolve and progressively elevate its reasoning process.

\section{Experiments}
\label{sec:exp}

\begin{table*}[t]
  \centering
  \caption{\textbf{Results on General Image Reasoning and Understanding Tasks.} We evaluate 10 benchmarks covering both visual reasoning and perception. Average scores are computed separately for the Reasoning and Perception groups with normalized scales for fair comparison. Insight-V significantly improves reasoning while maintaining strong perception, and Insight-V++ further advances performance on stronger base models.}
    \resizebox{\linewidth}{!}{
    \begin{tabular}{L{130pt}C{20pt}C{30pt}C{30pt}C{30pt}C{30pt}C{30pt}C{55pt}C{55pt}C{30pt}C{30pt}C{30pt}C{30pt}C{40pt}C{40pt}}
    \toprule
    \multirow{2}{*}{Model} & \multirow{2}{*}{Size} & \multicolumn{6}{c}{Reasoning} & \multicolumn{4}{c}{Perception} & \multicolumn{2}{c}{\cellcolor{blue!10}Average} \\
    \cmidrule(lr){3-8} \cmidrule(lr){9-12} \cmidrule(lr){13-14}
     &  & \rotatebox{45}{MMMU\textsubscript{val}} & \rotatebox{45}{MMMU-Pro\textsubscript{avg}} & \rotatebox{45}{MMBench} & \rotatebox{45}{ChartQA} & \rotatebox{45}{MMStar} & \rotatebox{45}{MathVista\textsubscript{testmini}} & \rotatebox{45}{MME} & \rotatebox{45}{TextVQA} & \rotatebox{45}{OCRBench} & \rotatebox{45}{AI2D} & \rotatebox{45}{Reasoning} & \rotatebox{45}{Perception} \\
    \midrule 
    DeepSeek-VL~\cite{lu2024deepseek} & 7B & 35.4 & - & 73.5 & 59.1 & 37.1 & 36.1 & -/- & - & - & 65.3 & -& - \\
     VILA-1.5~\cite{lin2023vila} & 8B & 38.6 & - & 75.3 & - & 39.7 & - & 1634.9/- & - & - & 58.8 & - & -\\
     Cambrian-1~\cite{tong2024cambrian} & 8B & 42.7 & - & 75.9 & 73.3 & - & 49.0 & 1547.1/- & 72.6 & - & 74.6 & - & -\\
    InternLM-XComposer2~\cite{dong2024internlm} & 7B & 41.1 & - & 77.6 & 71.8 & 56.2 & 59.5 & 2220.4 & 69.7 & - & 81.2 & - & -\\
    POINTS~\cite{liu2024points} & 7B & 51.4 & - & 78.0 & - & 60.9 & 63.0 & 2184.1 & 60.0 & - & 81.2 & - & - \\
    IXC-2.5~\cite{internlmxcomposer2_5} & 7B & 42.9 & - & 79.4 & 82.2 & 59.9 & 63.7 & 2233.1 & - & - & - & - & -\\
    Bunny-LLaMA3~\cite{he2024bunny} & 8B & 43.4 & - & 77.2 & - & - & 34.4 & 1588.9/321.1 & 62.1 & - & 69.4 & - & -\\
    MM-1.5~\cite{zhang2024mm1} & 7B & 41.8 & - & - & 78.6 & - & 47.6 & 1514.9/346.4 & - & - & - & - & -\\
    MiniCPM-LLaMA3-V 2.5~\cite{yao2024minicpmv} & 8B & 45.8  & 19.6 & 77.2 & - & 51.8 & 54.3 & 2024.6 & 725 & - & 78.4 & - & -\\
        MiniCPM-V-2.6~\cite{yao2024minicpm} & 7B & 49.8 & 27.2 & 78.0 & - & 57.5 & 60.6 & 2268.7 & 78.3 & 852 & 82.1 & - & 81.7\\
        Qwen2-VL~\cite{qwen2vl} & 7B & 53.7 & - & 81.0 & 83.0 & 60.7 & 61.4 & - & - & 877 & 83.0 & - & -\\
    Idefics3-LLaMA3~\cite{laurenccon2024building} & 8B & 46.6 & 22.9 & 77.5 & 74.8 & 55.9 & 58.4 & 1937.4 & 73.2 & - & 76.5 & 56.0 & -\\
    Ovis1.5-LLaMA3~\cite{lu2024ovis} & 8B & 48.3 & 23.6 & 76.6 & 76.4 & 57.3 & 63.0 & 1948.5 & 	
74.0 & - & 82.5 & 57.5 & -\\
    \midrule
    LLaVA-NeXT-LLaMA3~\cite{liu2024llavanext} & 8B & 36.9 & 13.2 & 72.3 & 69.4 & 43.1 & 45.9 & 1611.1/346.0 & 65.3 & 553 & 71.5 & 46.8 & 65.5\\
    + Multi-Agent & 8B & 40.8 & 17.8 & 77.6 & 74.6 & 52.6 & 47.4 & 1603.7/469.3 & 68.9 & 631 & 75.7 & 51.8 & 70.5\\
    \rowcolor{Gray} + Iterative DPO (\textcolor{Red}{\textbf{Insight-V-LLaVA}}) & 8B & 42.0 & 21.0 & 81.7 & 77.4 & 57.4 & 49.8 & 1583.9/485.4 & 70.5 & 663 & 77.3 & 54.9 & 72.3\\
    \midrule
    Our Base Model & 7B & 47.1 & 22.6 & 81.3 & 75.7 & 57.0 & 56.9 & 1573.7/482.5 & 75.4 & 713 & 79.7 & 56.8 & 75.0\\
    + Multi-Agent & 7B & 49.7 & 23.8 & 82.2 & 81.2 & 58.6 & 58.7 & 1662.2/629.3 & 77.0 & 738 & 80.1 & 59.0 & 78.4 \\
    \rowcolor{Gray} + Iterative DPO (\textcolor{Red}{\textbf{Insight-V}}) & 7B & 50.2 & 24.9 & 82.3 & 81.5 & 61.5 & 59.9 & 1685.1/627.0 & 76.8 & 735 & 79.8 & 60.1 & 78.0\\
    \midrule
    Qwen2.5-VL~\cite{bai2025qwen25vl} & 7B & 58.6 & 38.3 & 83.5 & 84.5 & 63.9 & 69.2 & 2347 & 81.2 & 842 & 82.4 & 66.3 & 82.9\\
    + Multi-Agent (RL) & 7B & 62.5 & 42.9 & 84.5 & 85.9 & 66.4 & 74.5 & 2415 & 80.4 & 822 & 81.4 & 69.5 & 82.6\\
    \rowcolor{Gray} + Self-Evolving (\textcolor{Red}{\textbf{Insight-V++}}) & 7B & 64.8 & 45.6 & 84.5 & 86.1 & 68.2 & 77.6 & 2406 & 80.6 & 830 & 81.7 & 71.1 & 82.8\\
    \bottomrule
    \end{tabular}%
    }
  \label{tab:reason}
\end{table*}%

We conduct extensive experiments across multiple vision-language benchmarks to validate the effectiveness of our method. In this section, we first introduce the implementation details of Insight-V in Section~\ref{sec:detail}. Then we present a comparison with state-of-the-art MLLMs, outlining the primary results of our method on image reasoning tasks as well as additional results on general image understanding in Section~\ref{sec:exp-visual}, and present the results on video understanding and reasoning in Section~\ref{sec:videoexp}. Moreover, we offer further analytical experiments and essential ablation studies on design choices in Section~\ref{sec:analysis}, along with qualitative results shown in Section~\ref{sec:qualitative} for more insights.

\subsection{Implementation Details}
\label{sec:detail}
\paragrapha{Data Construction Details.}
For the reasoning agent, to construct our image dataset, we utilize widely used open-source training datasets, including LLaVA-OneVision~\cite{li2024llavaov}, Cauldron~\cite{laurençon2024cauldron}, and Cambrian-1~\cite{tong2024cambrian}. Our compiled dataset is divided into two sections. The first section involves creating reasoning chains for challenging questions. We use Qwen2-VL 72B~\cite{qwen2vl} to identify questions it struggles with, which represent challenges for existing MLLMs, resulting in 100K data points. The second section concentrates on key aspects for MLLMs, such as OCR and math. We subsample data focusing on these aspects, ensuring it is distinct from the first section. We then filter out data that does not require complex reasoning (questions solvable in fewer than three steps), ultimately subsampling 100K data points. For Insight-V++, we further incorporate data from recent open-source works focusing on visual reasoning, refining their quality through our data generation pipeline. This process results in a total of approximately 600K image samples. For video data, we curate samples from several large-scale open-source datasets, including LLaVA-Video\cite{zhang2024llavavideo}, Video-R1\cite{feng2025videor1}, and Oryx~\cite{liu2024oryx}, bringing the combined dataset for supervised fine-tuning to around 1 million image–video pairs. For reinforcement learning training, we primarily collect image data from open-source datasets such as ViRL\cite{wang2025vlrethinker}, ThinkLite\cite{wang2025thinklitevl}, and OpenMMReasoner~\cite{zhang2025openmmreasoner}, while video data are mainly sourced from Video-R1~\cite{feng2025videor1} and the datasets used in supervised fine-tuning. To enhance data quality and training efficiency, we apply reject sampling based on pass@8 performance, retaining only challenging samples suitable for RL optimization. This results in approximately 120K high-quality data points for RL training.

For the summary agent, we initially combine data containing reasoning paths of varying quality with standard question–answer pairs. In Insight-V++, we further extend this dataset to the video question answering domain to preserve and enhance the model’s video perception capabilities. We further utilize Gemini-2.5-Pro~\cite{team2025gemini} to annotate the detailed flaws of the reasoning paths, which will help the summary agent to learn to identify the specific flaw patterns during visual reasoning. For reinforcement learning training, we leverage data generated during the annotation pipeline, including reasoning paths of different quality levels, thereby providing samples of varying difficulty for more effective and robust RL optimization.

\paragrapha{Model Training Details.}
We integrate the Insight-V system with various MLLMs to demonstrate the broad applicability of our approach. Our initial implementation with Insight-V on LLaVA-NeXT-LLaMA3~\cite{liu2024llavanext} illustrates the method's effectiveness. To further validate its generalizability and establish a solid baseline against state-of-the-art MLLMs, we additionally train a base multi-modal model using the Qwen-2.5-7B~\cite{qwen2.5} LLM. During pretraining, we utilize the 558K captioning dataset from LLaVA-1.5~\cite{liu2024llava15}, unfreezing the connector parameters. This is followed by supervised fine-tuning with a curated dataset of approximately 4 million images, using a learning rate of 2e-5 as guided by prior researches~\cite{liu2024llava15,liu2024llavanext}. This two-stage training process equips the baseline model with essential visual perception abilities, achieving competitive results on vision-language benchmarks. We then initialize two agents from the baseline model, performing targeted fine-tuning to obtain the final agents. For the reasoning agent, we compile a dataset of 200K images and train the model over 2 epochs with a learning rate of 5e-6. For the summary agent, we use a dataset of 1.2 million images, applying a learning rate of 1e-5 and training for 1 epoch. Additionally, we apply Direct Preference Optimization (DPO) to the reasoning agent, using approximately 15K preference data and training for 1 epoch at a learning rate of 5e-7. This DPO process is iteratively conducted across 3 rounds by using the model from the previous stage to generate preference data, thus enhancing the agent's reasoning capabilities. The lmms-eval~\cite{li2024xinrun} is utilized for fast evaluation.

For the further training of Insight-V++ and ST-GRPO/J-GRPO, we use up to 128 video frames as input. The learning rates are set to 1e-5 for SFT and 2e-6 for RL training, with a batch size of 128 for both stages. During RL training, the maximum output length is set to 16,384 tokens, with a temperature of 1.0 to maintain response diversity.

\begin{table*}[!t]
  \centering
  \caption{\textbf{Results on Advanced Image Reasoning Tasks.} We conduct evaluation experiments across 6 visual reasoning benchmarks, covering mathematical, logical, and chart-based advanced reasoning assessments. Insight-V++ achieves the highest average score, outperforming other strong RL-based methods. $\dagger$ means the results are reproduced by OpenMMReasoner~\cite{zhang2025openmmreasoner}.}
    \resizebox{\linewidth}{!}{
    \begin{tabular}{L{110pt}C{20pt}C{55pt}C{55pt}C{50pt}C{50pt}C{50pt}C{55pt}C{35pt}}
    \toprule
    Model & Size & MathVision\textsubscript{test} & MathVerse\textsubscript{testmini} & WeMath\textsubscript{Loose} & LogicVista\textsubscript{test} & DynaMath\textsubscript{worst} & CharXiv\textsubscript{Reasoning} & \cellcolor{blue!10}Average \\
    \midrule
    \multicolumn{9}{l}{\textit{Closed-source Models}} \\
    GPT-4o~\cite{hurst2024gpt4o} & - & 31.1 & 40.6 & - & 64.4 & 34.5 & - & - \\
    GPT-4o mini~\cite{hurst2024gpt4o} & - & 27.3 & 30.0 & 48.8 & 41.4 & 31.6 & 34.1 & - \\
    \midrule
    \multicolumn{9}{l}{\textit{Open-source Models -- SFT}} \\
    LLaVA-OneVision~\cite{li2024llavaov} & 7B & 17.6 & 17.6 & - & 32.0 & 9.0 & 23.6 & - \\
    InternVL3-8B~\cite{zhu2025internvl3} & 8B & 28.6 & 33.9 & - & 43.6 & 23.0 & 37.6 & - \\
    LLaVA-OneVision-1.5~\cite{li2024llavaov} & 8B & 25.6 & 46.3 & 49.4$^\dagger$ & 45.8$^\dagger$ & 19.8$^\dagger$ & 37.0$^\dagger$ & 37.3 \\
    \midrule
    \multicolumn{9}{l}{\textit{Open-source Models -- RL-based}} \\
    VLAA-Thinker-Qwen2.5~\cite{chen2025vlaathinker} & 7B & 26.4 & 48.2 & - & 48.5 & 22.4 & - & - \\
    ThinkLite-VL~\cite{wang2025thinklitevl} & 7B & 24.6 & 42.9 & - & 42.7 & 16.5 & - & - \\
    VL-Rethinker~\cite{wang2025vlrethinker} & 7B & 28.4 & 46.4 & - & 42.7 & 17.8 & - & - \\
    % M2-Reasoning~\cite{chen2025m2reasoning} & 7B & - & 40.4 & - & 50.6 & - & - & - \\
    MMR1~\cite{peng2025lmmr1} & 7B & 31.8 & 55.4 & 68.0$^\dagger$ & 48.9 & 27.9$^\dagger$ & 43.5$^\dagger$ & 45.9 \\
    OpenVLThinker~\cite{deng2025openvlthinker} & 7B & 23.0 & 38.1 & 61.9$^\dagger$ & 44.5 & 16.8 & 41.0$^\dagger$ & 37.6 \\
    MM-Eureka-Qwen~\cite{meng2025mmeureka} & 7B & 28.1 & 45.4 & 59.8$^\dagger$ & 46.3 & 23.0 & 42.4 & 40.8 \\
    Open-Vision-Reasoner~\cite{wei2025ovr} & 7B & 51.8 & 54.6 & 64.8 & 54.8 & 33.5 & 44.5 & 50.7 \\
    OpenMMReasoner~\cite{zhang2025openmmreasoner} & 7B & 43.6 & 63.8 & 79.0 & 50.0 & 34.9 & 46.1 & 52.9 \\
    \midrule
    Qwen2.5-VL~\cite{bai2025qwen25vl} & 7B & 25.5 & 41.1 & 53.1 & 47.9 & 21.8 & 36.4 & 37.6 \\
    + Multi-Agent (RL) & 7B & 44.8 & 59.1 & 77.6 & 51.7 & 31.4 & 44.5 &  51.5 \\
    \rowcolor{Gray} + Self-Evolving (\textcolor{Red}{\textbf{Insight-V++}}) & 7B & 48.6 & 62.4 & 78.8 & 52.9 & 33.6 & 46.8 &  53.9 \\
    \bottomrule
    \end{tabular}%
    }
  \label{tab:adv_reason}
\end{table*}%

\subsection{Results on Image Reasoning and Perception}
\label{sec:exp-visual}

\paragrapha{Setup.} We conduct comprehensive experiments across a diverse set of benchmarks to evaluate our models’ visual reasoning and perception capabilities. 
The evaluation is organized into two main categories: (\textit{i}) \textbf{general image reasoning and perception}, which assess foundational reasoning and visual understanding skills, and (\textit{ii}) \textbf{advanced image reasoning}, which targets complex multimodal, mathematical, and logical reasoning abilities. 
For the {general reasoning and perception benchmarks} in Table~\ref{tab:reason}, 
we include ten widely used benchmarks that jointly measure reasoning depth and perceptual accuracy.  MMMU~\cite{yue2024mmmu} and MMMU-Pro~\cite{yue2024mmmupro} evaluate expert-level perception and reasoning across diverse domains. MMBench~\cite{liu2023mmbench} offers a comprehensive assessment of multimodal large language models (MLLMs) in image-based understanding. ChartQA~\cite{masry2022chartqa} focuses on logical reasoning over graphical data, while MathVista~\cite{lu2023mathvista} examines mathematical problem‑solving abilities, following the standard GPT‑4‑0613~\cite{OpenAI_GPT4_2023} evaluation protocol. MMStar~\cite{chen2024we} further provides balanced coverage across multiple levels of task difficulty. To verify that our models preserve perceptual competence while enhancing reasoning, we also include vision centric benchmarks like MME~\cite{fu2023mme}, which comprises 14 subtasks evaluating perception and cognitive grounding, as well as TextVQA~\cite{singh2019textvqa}, OCRBench~\cite{liu2023ocrbench}, and AI2D~\cite{kembhavi2016ai2d}, which emphasize fundamental visual comprehension rather than abstract reasoning.
For the {advanced image reasoning benchmarks}, we adopt a set of recently proposed tasks designed to push the limits of multimodal reasoning. Specifically, MathVision~\cite{wang2024measuring}, MathVerse~\cite{zhang2024mathverse}, WeMath~\cite{qiao2025we}, and DynaMath~\cite{zou2024dynamath} target visual mathematical reasoning; LogicVista~\cite{xiao2024logicvista} evaluates multimodal logical reasoning; and Charxiv~\cite{wang2024charxiv} focuses on complex scientific chart interpretation. 
We further compare our {Insight-V++} with recent state-of-the-art multimodal reasoning frameworks, including OpenMMReasoner~\cite{zhang2025openmmreasoner} and OpenVisionReasoner~\cite{wei2025ovr}, to comprehensively assess its advanced visual reasoning capabilities.

\paragrapha{Results on General Image Reasoning and Understanding Tasks.} As shown in Table~\ref{tab:reason}, Insight‑V demonstrates consistent and significant improvements across diverse general image reasoning and understanding benchmarks. When integrated with LLaVA‑NeXT, Insight‑V elevates the average reasoning performance from 46.8\% to 54.9\% and enhances perception from 65.5\% to 72.3\%, highlighting its effectiveness in reinforcing both reasoning depth and perceptual quality. In particular, substantial gains are observed on ChartQA (+8.0\%), MMStar (+14.3\%), and MMMU‑Pro (+7.8\%), reflecting the framework’s capability to handle tasks involving structured and multimodal reasoning. Performance also improves notably on MMMU (+5.1\%) and MMBench (+9.4\%), further validating its robustness across a wide range of visual reasoning and understanding tasks. To assess scalability, we extend the framework to the more advanced Qwen2.5‑VL backbone. Despite the model’s strong capability, Insight‑V++ delivers additional advances, raising the average reasoning score from 66.3\% to 71.1\% while maintaining comparable perception performance. On complex reasoning benchmarks, Insight‑V++ achieves 64.8\% on MMMU and 45.6\% on MMMU‑Pro, exceeding both the reinforcement learning‑based multi‑agent baseline and prior state‑of‑the‑art methods. These consistent improvements across different model scales and task complexities underscore the efficacy of the proposed iterative and self‑evolving mechanisms. By enabling collaborative multi‑agent reasoning while maintaining strong perceptual accuracy, Insight‑V and Insight‑V++ achieve state‑of‑the‑art performance on general image reasoning and understanding tasks, demonstrating the adaptability and scalability of the proposed framework.

\paragrapha{Results on Advanced Image Reasoning Benchmarks.}
As shown in Table~\ref{tab:adv_reason}, Insight-V++ exhibits consistently strong performance compared with previous state-of-the-art visual reasoning models. It achieves the highest average score across all evaluated models and delivers outstanding results on both mathematical visual reasoning and scientific chart analysis, underscoring its advanced visual reasoning capability. Notably, the performance gains of self-evolving design are more pronounced on benchmarks that demand complex reasoning, with an average improvement of +2.4\%, compared to +1.6\% on general image reasoning benchmarks reported in Table~\ref{tab:reason}. These observations emphasize the effectiveness of the proposed self-evolving design, which particularly enhances performance on tasks requiring deeper reasoning, thereby demonstrating the superior advantage of the self-evolving multi-agent system.

\vspace{-15pt}
\subsection{Results on Video Reasoning}
\label{sec:videoexp}

%%%NEW%%%
\begin{table*}[!t]
\centering
\caption{\textbf{Results on Video Reasoning Tasks.} We conduct evaluation experiments across 6 video benchmarks, covering both general video understanding and task-specific video reasoning assessments. Insight-V++ demonstrates strong effectiveness and generalizability in video domain as well, achieving consistent improvements across all benchmarks and surpassing competitive open-source video MLLMs at the similar scale by a notable margin.}
\resizebox{\linewidth}{!}{
\begin{tabular}{L{120pt}C{30pt}C{50pt}C{50pt}C{45pt}C{50pt}C{55pt}C{45pt}C{50pt}}
\toprule
Model & Size & VideoMME & VideoMMMU & MMVU\textsubscript{MC} & VideoMMLU & VideoHolmes & Video-TT\textsubscript{MC} & \cellcolor{blue!10}Average \\
\midrule
\multicolumn{9}{l}{\textit{Proprietary Models}} \\
GPT-4o~\cite{GPT4o} & - & 71.9 & 61.2 & 75.4 & 49.4 & 42.0 & 46.6 & 57.8 \\
Gemini-1.5-Pro~\cite{team2024gemini} & - & 75.0 & 53.9 & - & - & 41.2 & 42.3 & - \\
Gemini-1.5-Flash~\cite{team2024gemini} & - & 70.3 & 49.8 & - & 43.6 & - & - & - \\
Gemini-2.5-Pro~\cite{team2025gemini} & - & 84.3 & 83.6 & - & - & 45.0 & - & - \\
\midrule
\multicolumn{9}{l}{\textit{Open-Sourced Video MLLMs}} \\
LLaVA-OneVision~\cite{li2024llavaov} & 7B & 58.2 & 33.9 & 49.2 & 34.0 & - & - & - \\
% Oryx~\cite{liu2024oryx} & 7B & 58.3 & - & - & - & - & - & - \\
Oryx-1.5~\cite{liu2024oryx} & 7B & 58.8 & - & - & - & - & 44.8 & - \\
LLaVA-Video~\cite{zhang2024llavavideo} & 7B & 63.3 & 36.1 & 48.8 & 21.5 & - & 41.8 & - \\
VideoLLaMA3~\cite{zhang2025videollama} & 7B & 66.2 & 47.0 & - & - & - & - & - \\
MiniCPM-V 2.6~\cite{yao2024minicpm} & 8B & 60.9 & - & - & 39.3 & - & - & - \\
InternVL2.5~\cite{chen2024expanding} & 8B & 64.2 & - & - & 39.5 & 23.6 & 44.7 & - \\
% NVILA~\cite{liu2024nvila} & 7B & 64.2 & - & - & - & - & - & - \\
VideoChat-Flash~\cite{li2024videochatflash} & 7B & 65.3 & 41.7 & - & 27.7 & - & - & - \\
Video-R1~\cite{feng2025videor1} & 7B & 61.4 & 52.4 & 63.8 & - & 36.5 & - & - \\
VideoChat-R1~\cite{li2025videochatr1} & 7B & 65.2 & 49.6 & - & - & 33.0 & 42.7 & - \\
\midrule
Qwen2.5-VL~\cite{bai2025qwen25vl} & 7B & 65.1 & 47.4 & 60.0 & 37.5 & 34.7 & 42.3 & 47.3 \\
+ Multi-Agent (RL) & 7B & 67.2 & 55.8 & 63.9 & 46.8 & 38.5 & 45.4 &  52.9\\
\rowcolor{Gray} + Self-Evolving (\textcolor{Red}{\textbf{Insight-V++}}) & 7B & 67.8  & 57.6 & 65.0 & 48.4 & 40.2 & 46.8 &  54.2\\
\bottomrule
\end{tabular}%
}
\label{tab:video_reason}
\end{table*}%

\paragrapha{Setup.} To assess whether Insight-V++ generalizes beyond static images, we conduct extensive experiments on video reasoning benchmarks that demand temporal understanding, domain expertise, and multi-step inference over dynamic visual content. We adopt six benchmarks spanning diverse evaluation dimensions. VideoMME~\cite{fu2024videomme} provides a comprehensive assessment of multi-modal video understanding; we report accuracy without subtitles. VideoMMMU~\cite{hu2025videommmu} and MMVU~\cite{zhao2025mmvu} target professional-level knowledge acquisition and expert-level understanding across disciplines such as medicine, engineering, and the humanities. VideoMMLU~\cite{song2025videommlu} specifically evaluates STEM lecture comprehension, requiring the model to recognize dynamically presented formulas and perform subject-specific reasoning. VideoHolmes~\cite{cheng2025videoholmes} challenges models with deductive reasoning chains over long-form video, while Video-TT~\cite{zhang2025videott} focuses on temporal and textual reasoning in instructional scenarios. During training, we support up to 128 frames to capture fine-grained temporal dynamics. Following Video-R1~\cite{feng2025videor1}, we sample 64 frames for evaluation. We compare with leading proprietary systems like GPT-4o, Gemini-1.5-Pro/Flash, Gemini-2.5-Pro, and competitive open-source video MLLMs at the similar scale.

\paragrapha{Main Results.} The experimental results in Table~\ref{tab:video_reason} confirm the effectiveness and generalizability of Insight-V++ on video reasoning tasks. Based on the Qwen2.5-VL model, the multi-agent pipeline with reinforcement learning raises the average score from 47.3\% to 52.9\%, with a 5.6\% absolute improvement on average; and the subsequent self-evolving stage yields a further 1.3\% gain, reaching 54.2\%. This cumulative improvement confirms the effectiveness of our two-stage progression, which mirrors the trend observed in image reasoning (Table~\ref{tab:reason}), validating that the decomposition of reasoning and summarization transfers effectively from the image to the video modality.
We also notice that the gains are obvious on benchmarks requiring domain-specific knowledge and structured inference. On VideoMMLU, which requires recognizing dynamically presented formulas and performing STEM-specific reasoning, Insight-V++ achieves a 10.9\% absolute improvement from 37.5\% to 48.4\%, nearly matching GPT-4o's performance (49.4\%). We attribute this to the multi-agent framework's ability to decompose complex lecture content into structured reasoning steps before producing a final answer. On VideoMMMU, which targets professional-level knowledge across disciplines, we observe an 10.2\% gain, notably exceeding all open-source counterparts including Video-R1 (52.4\%) and VideoChat-R1 (49.6\%). On MMVU, a 5.0\% improvement from 60.0\% to 65.0\% further validates the benefit of explicit reasoning for expert-level comprehension. Consistent improvements are also observed on benchmarks emphasizing deductive and temporal reasoning. On VideoHolmes, Insight-V++ improves by 5.5 
\% from 34.7\% to 40.2\%, clearly surpassing Video-R1 at 36.5\%, and VideoChat-R1 at 33.0\%; on Video-TT, it achieves 46.8\%, comparable to GPT-4o at 46.6\%, indicating that structured reasoning particularly benefits temporal and textual reasoning. Even on the more holistic VideoMME, a 2.7\% improvement from 65.1\% to 67.8\% suggests that enhanced reasoning does not compromise general video understanding. These results, together with findings in Section~\ref{sec:exp-visual}, demonstrate that the core principle of Insight-V++ holds robustly across both image and video modalities.

\subsection{Further Analysis}
\label{sec:analysis}
In this section, we present comprehensive experiments to validate the design choices of Insight-V, emphasizing our approach's key contributions. Additionally, we include a case study to further demonstrate the qualitative effectiveness of Insight-V.

\paragrapha{Effectiveness of Multi-agent System.} To evaluate the effectiveness of the multi-agent system, we comprehensively compare Insight-V with alternative design choices. We begin by reporting the performance of the summary agent without a reasoning process to underscore the importance of the reasoning agent. For a comprehensive comparison, we also restructure the collected data using a Chain-of-Thought (CoT) template and train a model to perform sequential reasoning and subsequently answer questions. Additionally, we enable a model that can handle multi-turn conversations, facilitating reasoning and summarization in a multi-turn format. We use MMMU, ChartQA, MathVista, and MMStar as representative benchmarks to assess the performance across different methods. 

\begin{table}[t]
  \centering
  \caption{\textbf{Ablations on the Design Choice of Insight-V Series.} The multi-agent design outperforms other configurations, highlighting the critical role of reasoning and summarization decomposition.}
    \resizebox{\linewidth}{!}{
    \begin{tabular}{l|ccccc}
    \toprule
    Model & MMMU & ChartQA & MathVista & MMStar & Avg \\
    \midrule
    Baseline & 47.1 & 75.7 & 56.9 & 57.0 & 59.2 \\
    Vanilla - Direct SFT & 47.0 & 79.2 & 57.6 & 58.4 & 60.6 \\
    Multi-Turn Supervised & 48.1 & 79.6 & 57.9 & 58.2 & 61.0 \\
    Summary Agent Only & 47.5 & 76.3 & 57.3 & 57.9 & 59.8 \\
    \rowcolor{Gray} Multi-Agent & 49.7 & 81.2 & 58.7 & 58.6 & 62.1 \\
    
    \bottomrule
    \end{tabular}%
    }
  \label{analysis:design}
\end{table}%

\begin{table}[t]
  \centering
  \caption{\textbf{Ablations on the DPO training strategy.} Iterative DPO progressively enhances the model's reasoning capabilities, leading to improved performance.}
    \resizebox{\linewidth}{!}{
    \begin{tabular}{l|ccccc}
    \toprule
    Model & MMMU & ChartQA & MathVista & MMStar & Avg \\
    \midrule
    Insight-V (Multi-Agent) & 49.7 & 81.2 & 58.7 & 58.6 & 62.1 \\
    + RLAIF & 49.5 & 81.4 & 59.1 & 59.2 & 62.3 \\
    + DPO & 50.8 & 80.8 & 59.3 & 59.9 & 62.7 \\
    \rowcolor{Gray} + Iterative DPO & 50.2 & 81.5 & 59.9 & 61.5 & 63.3 \\
    \bottomrule
    \end{tabular}%
    }
  \label{analysis:dpo}
\end{table}%

\begin{table}[t]
  \centering
  \caption{\textbf{Ablations on the design choices of Insight-V++.} The multi-agent system significantly benefits from the integration of the enhanced GRPO-based RL strategy and the proposed self-evolving mechanism.}
    \resizebox{\linewidth}{!}{
    \begin{tabular}{l|cccc}
    \toprule
    Model & MMMU & MathVision & VideoMME & VideoMMMU \\
    \midrule
    Insight-V++ (Multi-Agent) & 59.2 & 38.7 & 65.8 & 49.6\\
    + ST-GRPO & 61.7 & 42.9 & 66.7 & 52.0 \\
    + ST-GRPO/J-GRPO & 62.5 & 44.8 & 67.2 & 53.4 \\
    \midrule
    Collaborative Inference & 63.2 & 46.7 & 67.5 & 53.9 \\
    Self-Evolving & 64.8 & 48.6 & 67.8 & 55.8 \\
    \bottomrule
    \end{tabular}%
    }
  \label{analysis:insightv++}
\end{table}%

As shown in Table~\ref{analysis:design}, the results indicate that the multi-agent system plays a critical role in enhancing the system’s visual understanding capabilities. Using only the summary agent without a reasoning process results in limited improvements in reasoning tasks, as the model lacks the necessary reasoning framework for optimal performance. Training the model to perform chain-of-thoughts reasoning, which is denoted as Insight-V (Vanilla-Direct SFT), yields modest gains, as this approach does not sufficiently emphasize critical judgments within the reasoning process, merging reasoning and summarization into a single task. Furthermore, models trained with multi-turn conversations still produce sub-optimal results, underscoring the importance of a multi-agent system that separates reasoning and summarization.

\begin{figure}[!t]
    \centering
    \includegraphics[width=\linewidth]{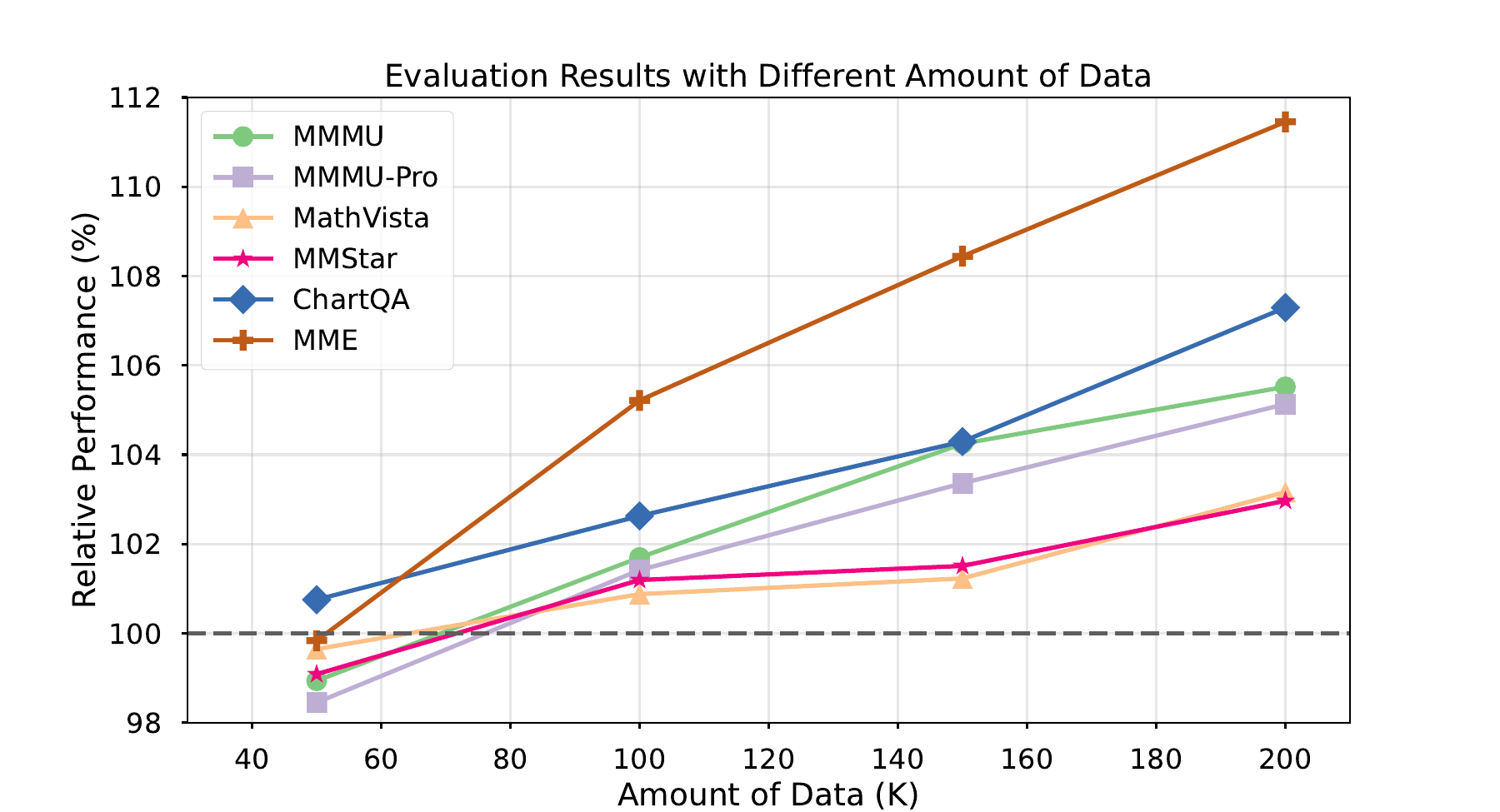} 
    \caption{\textbf{Ablations on the amount of training data.} The reasoning agent benefits from data scaling, providing more valuable insights for the summary agent.}
    \label{fig:scaling}
\end{figure}

\paragrapha{Data Scaling Law of Reasoning Agent.} To assess the effectiveness of the reasoning agent, we perform ablation experiments on the data volume used for training. As shown in Figure~\ref{fig:scaling}, we compare reasoning agents trained on varying data sizes: 50K, 100K, 150K, and 200K samples. For a fair and comprehensive comparison, we employ the same summary agent across evaluations and report results on six benchmarks. The findings clearly indicate that the reasoning agent benefits from increased data. With limited data, the reasoning agent struggles to generalize and fails to provide useful input to the summary model, resulting in performance even worse than baseline models. Conversely, training on larger datasets enhances the reasoning agent’s capabilities, enabling it to perform step-by-step reasoning and provide valuable insights that support the summary agent in solving tasks effectively.

\paragrapha{Effects of RL Algorithms. } We assess DPO algorithms to identify optimal alignment strategies for enhancing the reasoning process. To examine the effects of dataset composition on DPO training, we compare our curated dataset against the widely-used RLAIF-V~\cite{yu2024rlaif} dataset, which contains 80K DPO data pairs for alignment. We also investigate the potential benefits of iterative DPO in advancing the model's reasoning capabilities. For an unbiased comparison, we subsample the RLAIF-V dataset to approximately 15K preference data points, aligning it with the size of our dataset. The results, summarized in Table~\ref{analysis:dpo}, show an average improvement of approximately 0.2\% on the evaluated benchmarks when trained with the RLAIF-V dataset, whereas our curated dataset achieves greater gains of 0.6\%. This suggests that a DPO dataset based on model-generated rationales, rather than externally sourced data, more effectively boosts reasoning accuracy. Moreover, conducting two additional rounds of DPO training using the same methodology yields further performance gains of 0.6\%, underscoring that iterative DPO training progressively refines the model's ability to generate precise and high-quality reasoning processes compared to a single training pass.

\paragrapha{Design Choices of Insight-V++.} We conduct ablation studies to analyze the main design choices of Insight-V++, focusing on two key aspects. The first aspect involves the introduction of GRPO-based reinforcement learning algorithms. We ablate both the ST-GRPO and J-GRPO variants to assess their individual contributions. As shown in Table~\ref{analysis:insightv++}, ST-GRPO accounts for the largest performance gains, indicating that a strong reasoning agent is crucial for building a multi-agent system with advanced reasoning capabilities. Nevertheless, integrating J-GRPO into the summary agent further improves overall performance by enhancing its robustness and fundamental visual reasoning ability, thus leading to a stronger system overall. We also evaluate the effectiveness of the proposed self-evolving strategy. To this end, we compare it with a strong collaborative inference baseline, where the reasoning agent refines its reasoning path based on feedback from the summary agent, and the summary agent generates the final answer using the refined reasoning path. In contrast, our self-evolving strategy demonstrates superior performance, particularly on benchmarks requiring enhanced visual reasoning, such as MathVision and VideoMMMU. Unlike collaborative inference, where the two agents remain fixed after a single training round, the self-evolving mechanism enables continuous mutual improvement across multiple training iterations. This iterative refinement allows the multi-agent system to develop stronger reasoning abilities and achieve better results on more challenging benchmarks.

\begin{figure*}[t]
\centering
\includegraphics[width=0.98\textwidth]{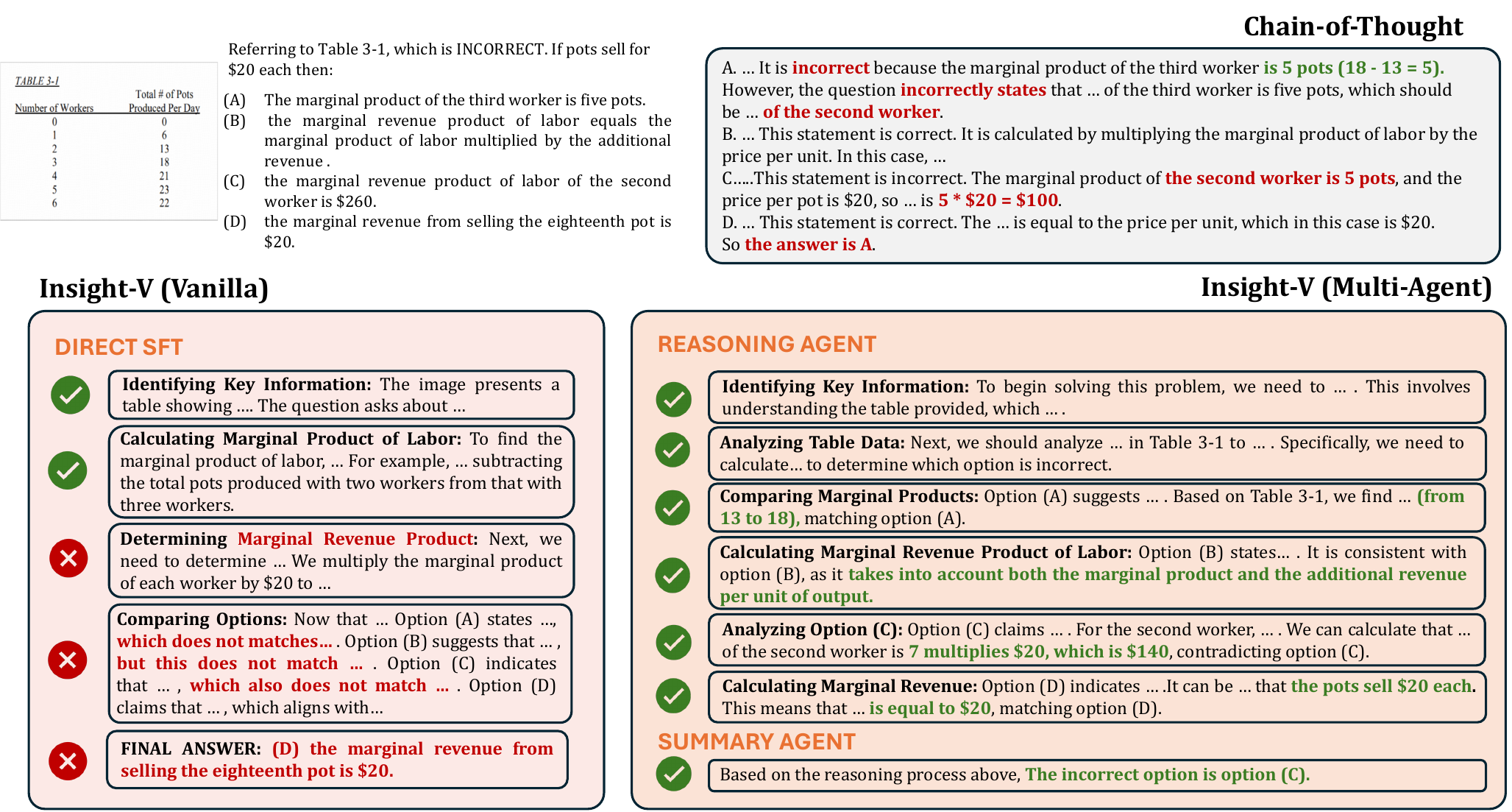}
\caption{\textbf{Qualitative Results of Insight-V Series.} We present qualitative comparisons of Insight-V with Chain-of-Thought and learning  Insight-V with direct SFT (Vanilla). For the Insight-V system, the reasoning agent delivers a more coherent and structured reasoning process, guiding the summary agent toward the correct answer, whereas other methods struggle with complex reasoning tasks and fail to solve such challenging problems. }
\label{fig:showcase}
\end{figure*}

\subsection{Qualitative Results}
\label{sec:qualitative}

We present qualitative comparisons in Figure~\ref{fig:showcase} to illustrate the improvements introduced by integrating the Insight-V system. Specifically, we provide an example that highlights advancements in the reasoning process. In this example, the model is challenged with a multiple-choice question based on a table. We compare our approach with a direct Chain-of-Thought application, and the model undergoes supervised fine-tuning without incorporating a multi-agent system, which is denoted as Insight-V (Vanilla). It is evident that applying Chain-of-Thought directly results in suboptimal reasoning and leads to incorrect answers. Although fine-tuning a single model yields somewhat better reasoning, the model begins to falter as the reasoning chain lengthens, ultimately arriving at incorrect answers. This outcome arises because the model is required to generate both reasoning steps and the final answer simultaneously, limiting its judgment capabilities and weakening its robustness in handling flawed reasoning chains.  In contrast, employing Insight-V enables more logical step decomposition and a structured reasoning chain. The model can analyze each option step-by-step and perform detailed calculations, which the other two methods cannot achieve. The fine-tuned summary model is able to evaluate the reasoning process and determine whether to derive the final answer based on it, significantly enhancing system robustness and ensuring correct answers.

\section{Conclusion}
\label{sec:conclusion}
In this paper, we present Insight-V series, a novel system that integrates a scalable data generation pipeline for producing long-chain, high-quality reasoning data with an efficient multi-agent training framework to enhance the reasoning capabilities of MLLMs. This system provides a scalable and effective approach to improving reasoning performance. To fully unlock the potential of the multi-agent collaborative reasoning framework, we further propose Insight-V++, which enhances the system with a GRPO-based reinforcement learning strategy and extends its application to video reasoning tasks. Moreover, we introduce a self-evolving architecture that enables agents to cooperate and co-evolve, leading to continual performance improvements. Extensive evaluations across diverse benchmarks demonstrate the effectiveness of our approach, paving the way toward MLLMs with stronger and more generalizable reasoning abilities.

\section*{Acknowledgement}
This study is supported by the Ministry of Education, Singapore, under its MOE AcRF Tier 2 (MOE-T2EP20221-0012, MOE-T2EP20223-0002), and under the RIE2020 Industry Alignment Fund – Industry Collaboration Projects (IAF-ICP) Funding Initiative, as well as cash and in-kind contributions from the industry partner(s).

\bibliographystyle{IEEEtranS}
\bibliography{main}

\end{document}